\documentclass{article}
\usepackage[final,nonatbib]{neurips_2024}
\usepackage[utf8]{inputenc} 

\usepackage[T1]{fontenc}    
\usepackage{mathptmx}
\usepackage{amsmath,amsfonts,amssymb,amsthm,graphicx}

\usepackage[normalem]{ulem}
\useunder{\uline}{\ul}{}

\usepackage{amsmath}
\usepackage{amssymb}
\usepackage{booktabs}
\usepackage{graphicx}
\usepackage{CJKutf8}
\usepackage{tabularx}
\usepackage{tikz,tcolorbox}
\usepackage{listings}
\usepackage{algorithm}
\usepackage{algpseudocode}
\usepackage{chngcntr}
\usepackage{multirow}
\usepackage[unicode]{hyperref}
\usepackage{xcolor}

\begin{document}

\footnotetext[1]{, *: These authors contributed equally to this work.}
\footnotetext[2]{, $\dagger$: Corresponding author.}

\title{KAG: Boosting LLMs in Professional Domains via Knowledge Augmented Generation} 

\author{\\Lei Liang$^{\ast,1}$, Mengshu Sun$^{\ast,1}$, Zhengke Gui$^{\ast,1}$, Zhongshu Zhu$^{1}$, Ling Zhong$^{1}$, Peilong Zhao$^{1}$, \\ 
Zhouyu Jiang$^{1}$, Yuan Qu$^{1}$, Zhongpu Bo$^{1}$, Jin Yang$^{1}$, 
Huaidong Xiong$^{1}$, Lin Yuan$^{1}$, Jun Xu$^{1}$, \\
Zaoyang Wang$^{1}$, Zhiqiang Zhang$^{1}$, Wen Zhang$^{2}$, Huajun Chen$^{2}$, Wenguang Chen$^{1}$, Jun Zhou$^{\dagger,1}$ \\
\\
\texttt{\{leywar.liang, mengshu.sms, zhengke.gzk, jun.zhoujun\}@antgroup.com} \\
\AND
$^{1}$Ant Group Knowledge Graph Team, $^{2}$Zhejiang University
\AND
{\texttt{Github:}\href{https://github.com/OpenSPG/KAG}{https://github.com/OpenSPG/KAG}}
}

\date{\today}

\setlength{\parindent}{0pt}

\maketitle

\begin{abstract}
The recently developed retrieval-augmented generation (RAG) technology has enabled the efficient construction of domain-specific applications. However, it also has limitations, including the gap between vector similarity and the relevance of knowledge reasoning, as well as insensitivity to knowledge logic, such as numerical values, temporal relations, expert rules, and others, which hinder the effectiveness of professional knowledge services. In this work, we introduce a professional domain knowledge service framework called Knowledge Augmented Generation (\textbf{KAG}). KAG is designed to address the aforementioned challenges with the motivation of making full use of the advantages of knowledge graph(KG) and vector retrieval, and to improve generation and reasoning performance by bidirectionally enhancing large language models (LLMs) and KGs through five key aspects: (1) LLM-friendly knowledge representation, (2) mutual-indexing between knowledge graphs and original chunks, (3) logical-form-guided hybrid reasoning engine, (4) knowledge alignment with semantic reasoning, and (5) model capability enhancement for KAG. We compared KAG with existing RAG methods in multi-hop question answering and found that it significantly outperforms state-of-the-art methods, achieving a relative improvement of 19.6\% on hotpotQA and 33.5\% on 2wiki in terms of F1 score. We have successfully applied KAG to two professional knowledge Q\&A tasks of Ant Group, including E-Government Q\&A and E-Health Q\&A, achieving significant improvement in professionalism compared to RAG methods. Furthermore, we will soon natively support KAG on the open-source KG engine OpenSPG, allowing developers to more easily build rigorous knowledge decision-making or convenient information retrieval services. This will facilitate the localized development of KAG, enabling developers to build domain knowledge services with higher accuracy and efficiency.
\end{abstract} 

\section{Introduction} 
Recently, the rapidly advancing Retrieval-Augmented Generation (RAG)\cite{gao2023retrieval,shao2023enhancing,chen2024benchmarking,fan2024survey,yu2023chain}  technology has been instrumental in equipping Large Language Models (LLMs) with the capability to acquire domain-specific knowledge. This is achieved by leveraging external retrieval systems, thereby significantly reducing the occurrence of answer hallucinations and allows for the efficient construction of applications in specific domains. In order to enhance the performance of the RAG system in multi-hop and cross-paragraph tasks, knowledge graph, renowned for strong reasoning capabilities, have been introduced into the RAG technical framework, including GraphRAG\cite{graphrag}, DALK\cite{li2024dalk}, SUGRE\cite{sugre}, ToG 2.0\cite{ma2024think}, GRAG\cite{grag}, GNN-RAG \cite{gnnrag} and HippoRAG\cite{gutierrez2024hipporag}.

Although RAG and its optimization have solved most of the hallucination problems caused by a lack of domain-specific knowledge and real-time updated information, the generated text still lacks coherence and logic, rendering it incapable of producing correct and valuable answers, particularly in specialized domains such as law, medicine, and science where analytical reasoning is crucial. This shortcoming can be attributed to three primary reasons.  Firstly, real-world business processes typically necessitate inferential reasoning based on the specific relationships between pieces of knowledge to gather information pertinent to answering a question. RAG, however, commonly relies on the similarity of text or vectors for retrieving reference information, which may lead to incomplete and repeated search results. secondly, real-world processes often involve logical or numerical reasoning, such as determining whether a set of data increases or decreases in a time series, and the next token prediction mechanism used by language models is still somewhat weak in handling such problems.

In contrast, the technical methodologies of knowledge graphs can be employed to address these issues. Firstly, KG organize information using explicit semantics; the fundamental knowledge units  are SPO triples, comprising entities and the relationships between them\cite{Paulheim2016KnowledgeGR}. Entities possess clear entity types, as well as relationships. Entities with the same meaning but expressed differently can be unified through entity normalization, thereby reducing redundancy and enhancing the interconnectedness of knowledge \cite{nodeclustering}. During retrieval, the use of query syntax (such as SPARQL\cite{nl2sparql} and SQL\cite{nl2sql}) enables the explicit specification of entity types, mitigating noisy from same named or similar entities, and allows for inferential knowledge retrieval by specifying relationships based on query requirements, as opposed to aimlessly expanding into similar yet crucial neighboring content. Meanwhile, since the query results from knowledge graphs have explicit semantics, they can be used as variables with specific meanings. This enables further utilization of the LLM's planning and function calling capabilities \cite{react}, where the retrieval results are substituted as variables into function parameters to complete deterministic inferences such as numerical computations and set operations.

To address the above challenges and meet the requirements of professional domain knowledge services, we propose \textbf{Knowledge Augmented Generation(KAG)}, which fully leverages the complementary characteristics of KG and RAG techniques. More than merely integrating graph structures into the knowledge base process, it incorporates the semantic types and relationships of knowledge graph and the commonly used Logical Forms from KGQA (Knowledge Graph Question Answering) into the retrieval and generation process. As shown in Figure \ref{fig:kag_framework}, this framework involves the optimization of the following five modules:

   \begin{itemize}
       \item \textbf{We proposed a LLM friendly knowledge representation framework LLMFriSPG}. We refer to the hierarchical structure of data, information, and knowledge of DIKW to upgrade SPG to be friendly to LLMs, named LLMFriSPG, to make it compatible with schema-free information extraction and schema-constrained expert knowledge construction on the same knowledge type (such as entity type, event type), and supports the mutual-indexing representation between graph structure and original text chunks, which facilitates the construction of graph-structure-based inverted index and facilitates the unified representation, reasoning, and retrieval of logical form.
       \item \textbf{We proposed a logical-form-guided hybrid solving and reasoning engine}. It includes three types of operators: \textit{planning, reasoning} and \textit{retrieval}, transforming natural language questions into a problem-solving process that combines language and symbols. Each step in the process can utilize different operators such as exact match retrieval, text retrieval, numerical computation, or semantic reasoning, thereby achieving the integration of four distinct problem-solving processes: retrieval, KG reasoning, language reasoning, and numerical computation.
       \item \textbf{We proposed a knowledge alignment approach based on semantic reasoning}. Define domain knowledge as various semantic relations such as \textit{synonyms, hypernyms}, and \textit{inclusions}. Semantic reasoning is performed in both offline KG indexing and online retrieval phases, allowing fragmented knowledge generated through automation to be aligned and connected through domain knowledge. In the offline indexing phase, it can improve the standardization and connectivity of knowledge, and in the online Q\&A phase, it can serve as a bridge between user questions and indexing accurately.
       \item \textbf{We proposed a model for KAG}. To support the capabilities required for the operation of the KAG framework, such as index construction, retrieval, question understanding, semantic reasoning, and summarization, we enhance the three specific abilities of general LLMs: Natural Language Understanding (NLU), Natural Language Inference (NLI), and Natural Language Generation (NLG) to achieve better performance in each functional module.
   \end{itemize}
We evaluated the effectiveness of the system on three complex Q\&A datasets: 2WikiMultiHopQA\cite{2wiki}, MuSiQue\cite{musique} and HotpotQA\cite{hotpotqa}. The evaluation focused on both end-to-end Q\&A performance and retrieval effectiveness. Experimental results showed that compared to HippoRAG\cite{gutierrez2024hipporag}, KAG achieved significant improvements across all three tasks, with F1 scores increasing by 19.6\%, 12.2\% and 12.5\% respectively. Furthermore, retrieval metrics also showed notable enhancements.

KAG is applied in two professional Q\&A scenarios within Ant Group: E-Government and E-Health. In the E-Government scenario, it answers users' questions about administrative processes based on a given repository of documents. For E-Health, it responds to inquiries related to diseases, symptoms, treatments, utilizing the provided medical resources. Practical application results indicate that KAG achieves significantly higher accuracy than traditional RAG methods, thereby enhancing the credibility of Q\&A applications in professional fields. We will soon natively support KAG on the open source KG engine OpenSPG, allowing developers to more easily build rigorous knowledge decision-making or convenient information retrieval services. 

In summary, we propose a knowledge-augmented technical framework, KAG, targeting professional question-answering scenarios and validate the effectiveness of this framework based on complex question-answering tasks. We present two industry application cases based on Ant Group's business scenarios and have open-sourced the code to assist developers in building local applications using KAG.

\section{Approach}
\begin{figure}[htbp]
    \centering
    \includegraphics[width=0.9\linewidth]{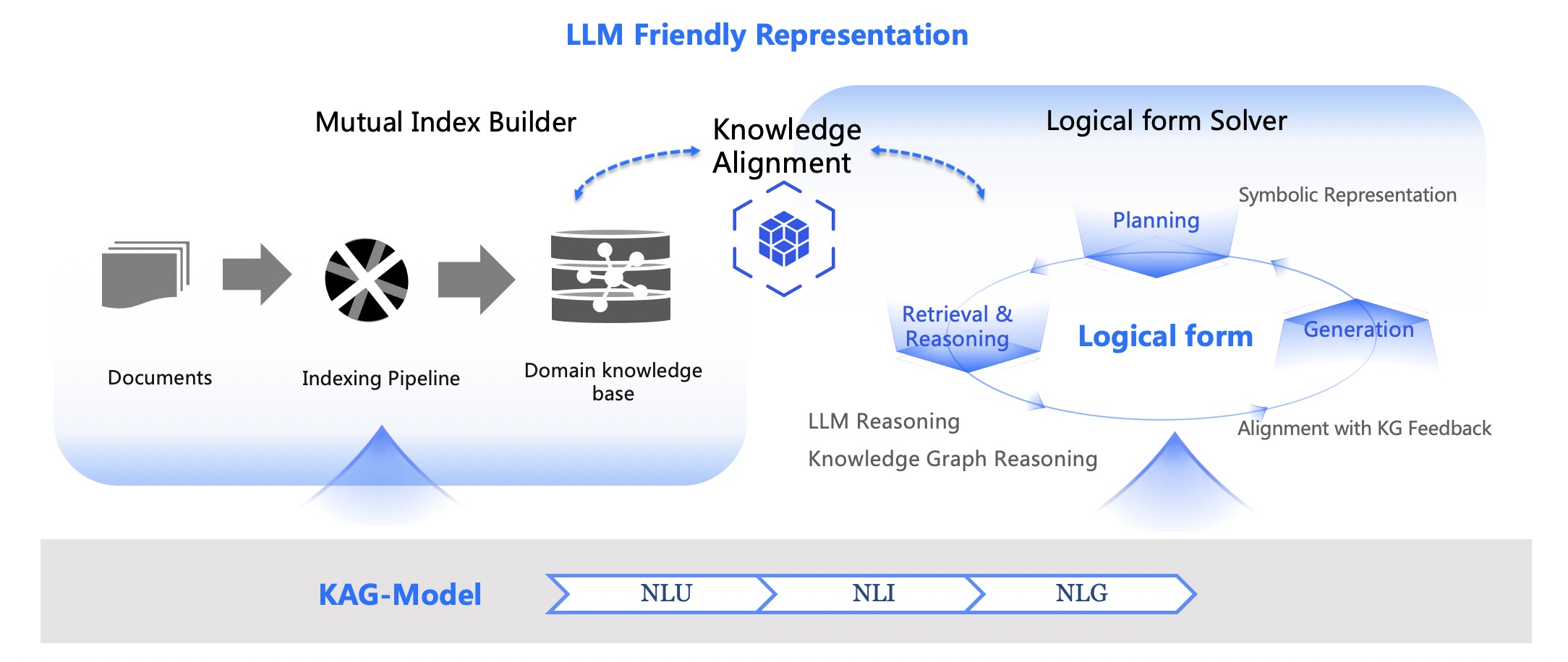}
    \caption{The KAG Framework. The left side shows KAG-Builder, while the right side displays KAG-Solver. The gray area at the bottom of the image represents KAG-Model. }
    \label{fig:kag_framework}
\end{figure}

In this section, we will first introduce the overall framework of KAG, and then discuss five key enhancements in sections 2.1 to 2.5. As shown in Figure \ref{fig:kag_framework}, the KAG framework consists of three parts: KAG-Builder, KAG-Solver, and KAG-Model. The KAG-Builder is designed for building offline indexes, in this module, we proposed a \textit{LLM Friendly Knowledge Representation framework} and \textit{mutual-indexing between knowledge structure and text chunk}. In the module KAG-Solver we introduced a \textit{Logical-form-guided hybrid reasoning solver} that integrates LLM reasoning, knowledge reasoning, and mathematical logic reasoning. Additionally, \textit{knowledge alignment by semantic reasoning} is used to enhance the accuracy of knowledge representation and retrieval in both KAG-Builder and KAG-Solver. The KAG-Model optimizes the capabilities needed by each module based on a general language model, thereby improving the performance of all modules.

\subsection{LLM Friendly Knowledge Representation}
\begin{figure}[htbp]
    \centering
    \includegraphics[width=0.8\linewidth]{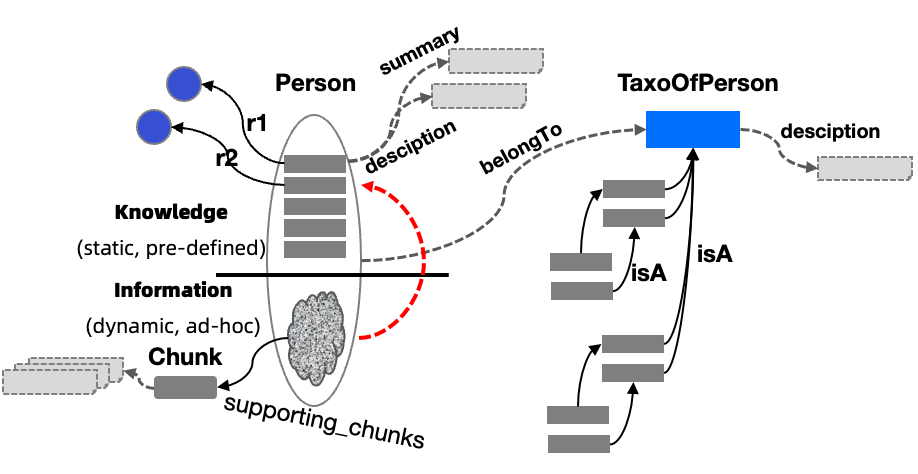}
    \caption{LLMFriSPG:A knowledge representation framework that is friendly to LLMs. Instances and concepts are separated to achieve more effective alignment with LLMs through concepts. In this study, entity instances and event instances are collectively referred to as instances unless otherwise specified. SPG properties are divided into knowledge and information areas, also called static and dynamic area, which are compatible with decision-making expertise with strong schema constraints and document retrieval index knowledge with open information representation. The red dotted line represents the fusion and mining process from information to knowledge. The enhanced document chunk representation provides traceable and interpretable text context for LLMs.}
    \label{fig:llmspg}
\end{figure}
In order to define a more friendly knowledge semantic representation for LLMs, we upgrade SPG from three aspects: deep text-context awareness, dynamic properties and knowledge stratification, and name it \textbf{LLMFriSPG}.

\begin{center}
$\mathcal{M}$ $=$ $\{\mathcal{T,\rho, C, L}\}$    
\end{center}
where, $\mathcal{M}$ represents all types defined in LLMFriSPG, $\mathcal{T}$ represents all \textbf{EntityType}(e.g., Person in Figure \ref{fig:llmspg}), \textbf{EventType} classes and all pre-defined properties that are compatible with LPG syntax declarations. $\mathcal{C}$ represents all \textbf{ConceptType} classes, concepts and concept relations, it is worth noting that the root node of each concept tree is a \textbf{ConceptType} class that is compatible with LPG syntax(e.g., TaxoOfPerson in Figure \ref{fig:llmspg}.), each concept node has a unique \textbf{ConceptType} class. $\mathcal{\rho}$ represents the inductive relations from instances to conecepts. $\mathcal{L}$ represents all executable rules defined on logical relations and logical concepts. For $\mathcal{\forall} {t} \in \mathcal{T}$:
\begin{center}
${p_t} = \{p_{t}^{c}, {p_t}^{f}, p_{t}^{b}\}$
\end{center}
As is show in Figure \ref{fig:llmspg}, where, $p_{t}$ represents all properties and relations of type $t$, and $p_{t}^{c}$ represents the domain experts pre-defined part, $p_{t}^{f}$ represents the part added in an ad-hoc manner, $p_{t}^{b}$ represents the system built-in properties, such as \textit{supporting\_chunks, descripiton, summary} and \textit{belongTo}. 
 For any instance $e_{i}$, denote $typeof(e_{i})$ as $t_{k}$, and \textit{supporting\_chunks} represents the set of all text chunks containing instance ${e_{i}}$, the user defines the chunk generation strategy and the maximum length of the chunk in KAG builder phase, \textit{description} represents the general descriptive information specific to class ${t_{k}}$. It is worth noting that the meaning of \textit{description} added to the type ${t_{k}}$ and the instance ${e_{i}}$ is different, when \textit{description} is attached to ${t_{k}}$, it signifies the global description for that type. Conversely, when it is associated with an instance ${e_{i}}$, it represents the general descriptive information for ${e_{i}}$ consistent with the orignal document context, \textit{description} can effectively assist LLM in understanding the precise meaning of a specific instance or type, and can be used in tasks such as information extraction, entity linking, and summary generation. \textit{summary} represents the summary of ${e_{i}}$ or ${r_{j}}$ in the original document context. \textit{belongTo} represents the inductive semantics from instance to concept. Each \textbf{EntityType} or \textbf{EventType} can be associated with a \textbf{ConceptType} through \textit{belongTo}. It is worth noting that, \textbf{1)} $\mathcal{T}$ and $\mathcal{C}$ \textbf{have different functions.} The statement ${t}$ adopts the object-oriented principle to better match the representation of the LPG\cite{sharma2019schema}, and $\mathcal{C}$ is managed by a text-based concept tree. This article will not introduce the SPG semantics in detail. \textbf{2) $p_{t}^{c}$ and $p_{t}^{f}$ can be instantiated separately}. That is, they share the same class declaration, but in the instance storage space, pre-defined static properties and realtime-added dynamic properties can coexist, and we also support instantiating only one of them. This approach can better balance the application scenarios of professional decision-making and information retrieval. General information retrieval scenarios mainly instantiate dynamic properties, while professional decision-making application scenarios mainly instantiate static properties. Users can strike a balance between ease of use and professionalism based on business scenario requirements. \textbf{3) $p_{t}^{c}$ and $p_{t}^{f}$ share the same conceptual terminology}. Concepts are general common sense knowledge that is independent of specific documents or instances. Different instances are linked to the same concept node to achieve the purpose of classifying the instances. We can achieve semantic alignment between LLM and  instances through concept graphs, and concepts can also be used as navigation for knowledge retrieval. the details are shown in section 2.4 and 2.3. \newline
 \begin{figure}[htbp]
     \centering
     \includegraphics[width=1\linewidth]{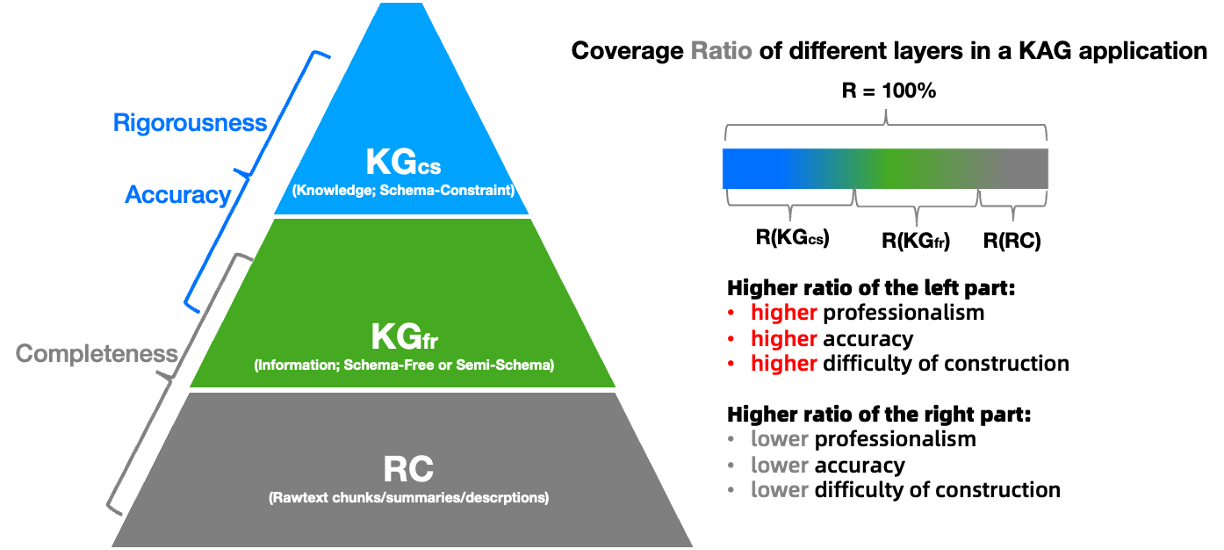}
     \caption{Hierarchical representation of knowledge and information. }
     \label{fig:hie_know_info}
 \end{figure}

 In order to more accurately define the hierarchical representation of information and knowledge, as shown in \ref{fig:hie_know_info}, we denote $KG_{cs}$ as \textit{knowledge layer}, which represents the domain knowledge that complies with the domain schema constraints and has been summarized, integrated, and evaluated. denote $KG_{fr}$ as \textit{graph information layer}, which represents the graph data such as entities and relations obtained through information extraction. denote $RC$ as \textit{raw chunks layer}, which represents the original document chunks after semantic segmentation. the $KG_{cs}$ layer fully complies with the SPG semantic specification and supports knowledge construction and logical rule definition with strict schema constraints, SPG requires that domain knowledge must have pre-defined schema constraints. It has high knowledge accuracy and logical rigor. However, due to its heavy reliance on manual annotation, the labor cost of construction is relatively high and the information completeness is insufficient. $KG_{fr}$ shares the same EntityTypes, Eventtypes and Conceptual system with $KG_{cs}$, and provides effective information supplement for $KG_{cs}$. Meanwhile, the \textit{supporting\_chunks, summary}, and \textit{description} edges built between $KG_{fr}$ and $RC$ form an inverted index based on graph structure, making $RC$ an effective original-text-context supplement for $KG_{fr}$ and with high information completeness. As is show in the right part of figure \ref{fig:hie_know_info}, in a specific domain application, $R(KG_{cs})$, $R(KG_{fr})$, and $R(RC)$ respectively represent their knowledge coverage in solving the target domain problems. If the application has higher requirements for knowledge accuracy and logic rigorousness, it is necessary to 
 build more domain structured knowledge and consume more expert manpower to increase the coverage of $R(KG_{cs})$. On the contrary, if the application has higher requirements for retrieval efficiency and a certain degree of information loss or error tolerance, it is necessary to increase the coverage of $R(KG_{fr})$ to fully utilize KAG's automated knowledge construction capabilities and reduce expert manpower consumption.

\begin{figure}[htbp]
    \centering
    \includegraphics[width=1\linewidth]{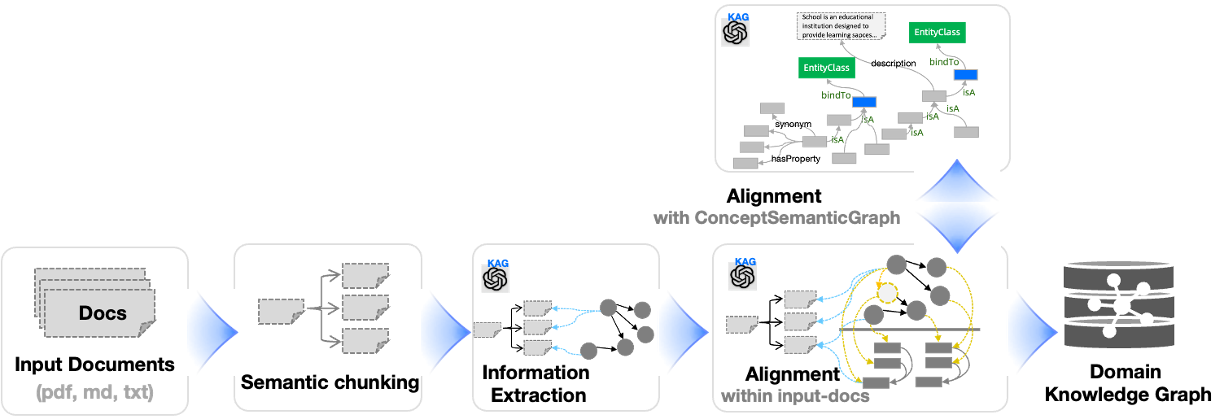}
    \caption{The Pipeline of KAG Builder for domain unstructured documents. From left to right, first, phrases and triples are obtained through information extraction, then disambiguation and fusion are completed through semantic alignment, and finally, the constructed KG is written into the storage.}
    \label{fig:kg_builder}
\end{figure}

\subsection{Mutual Indexing}
As illustrated in Figure \ref{fig:kg_builder}, KAG-Builder consists of three coherent processes: structured information acquisition, knowledge semantic alignment and graph storage writer. The main goals of this module include: 1) building a mutual-indexing between the graph structure and the text chunk to add more descriptive context to the graph structure, 2) using the concept semantic graph to align different knowledge granularities to reduce noise and increase graph connectivity.

\subsubsection{Semantic Chunking}

According to the document's structural hierarchy and the inherent logical connections between paragraphs, a semantic chunking process is implemented based on system-built-in prompts. This semantic chunking produces chunks that adhere to both length constraints (specifically for LLM's context window size constraints) and semantic coherence, ensuring that the content within each chunk is thematically cohesive. We defined \textbf{Chunk EntityType} in $RC$, which includes fields such as \textit{id, summary, and mainText}. Each chunk obtained after semantic segmentation will be written into an instance of \textbf{Chunk}, where \textit{id} is a composite field consisting of \textit{articleID, paraCode, idInPara} concatenated by the connector \textit{\#} in order to ensure that consecutive chunks are adjacent in the \textit{id} space. \textit{articleID} represents the globally unique article ID, \textit{paraCode} represents the paragraph code in the article, and \textit{idInPara} is the sequential code of each chunk in the paragraph. Consequently, an adjacency in the content corresponds to a sequential adjacency in their identifiers. Furthermore, a reciprocal relation is established and maintained between the original document and its segmented chunks, facilitating navigation and contextual understanding across different granularities of the document's content. This structured approach to segmentation not only optimizes compatibility with large-scale language models but also preserves and enhances the document's inherent semantic structure and association.
\begin{figure}[htbp]
    \centering
    \includegraphics[width=0.9\linewidth]{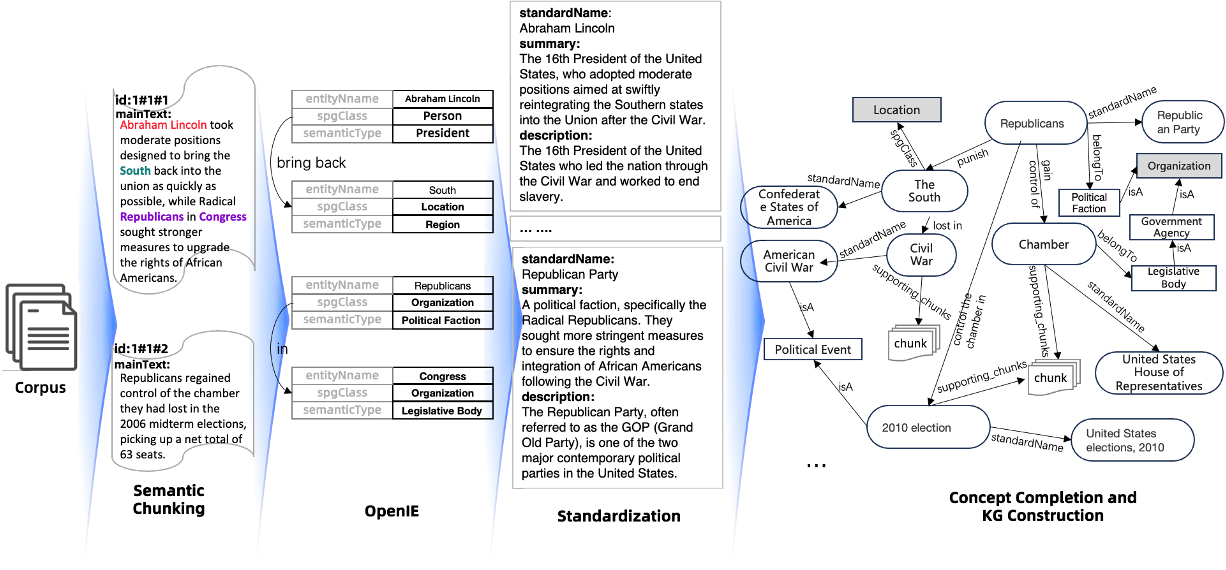}
    \caption{An Example of KAG-Builder pipeline}
    \label{fig:kag-builder-example}
\end{figure}
\subsubsection{Information Extraction with More Descriptive Context}
Given a dataset, we use fine-tuning-free LLM(such as GPT-3.5, DeepSeek, QWen, etc,.) or our fine-tuned model Hum to extract entities, events, concepts and relations to construct $KG_{fr}$, subsequently, construct the mutual-indexing structure between $KG_{fr}$ and $RC$, enabling cross-document links through entities and relations. This process includes three steps. First, it extracts the entity set $E = \{e_1, e_2, e_3, ...\}$ chunk by chunk, second, extracts the event set ${EV} = \{{ev}_1, {ev}_2, {ev}_3, ...\}$ associated to all entities and iteratively extracts the relation set $R = \{r_1, r_2, r_3, ...\}$ between all entities in $E$, finally, completes all hypernym relations between the instance and its \textit{spgClass}. To provide more convenience for the subsequent Knowledge Alignment phase, and overcome the problem of low discrimination of knowledge phrases such as Wikidata\cite{vrandevcic2014wikidata} and ConceptNet\cite{liu2004conceptnet}, in the entity extraction phase, we use LLMs to generate built-in properties \textit{description, summary, semanticType, spgClass, descripitonOfSemanticType} by default for each instance \textit{e} at one time, as shown in Figure \ref{fig:llmspg}, we store them in the \textit{e} instance storage according to the structure of \textit{e.description,e.summary, <e, belongTo, semanticType> and <e, hasClass, spgClass>}. 
\subsubsection{Domain Knowledge Injection And Constraints}

When openIE is applied to professional domains, irrelevant noise will be introduced. Previous researches\cite{chen2024benchmarking,yu2023chain,wu2024easily} have shown that noisy and irrelevant corpora can significantly undermine the performance of LLMs. It is a challenge to align the granularity of extracted information and domain knowledge. The domain knowledge alignment capabilities in KAG include: \textbf{1) Domain term and concept injection}. We use an iterative extraction approach, First, we store domain concepts and terms with \textit{description} in KG storage. Second, we extract all instances in the document through openIE, then we perform vector retrieval to obtain all possible concept and term sets $E_d$. Finally, we add $E_d$ to the extraction prompt and perform another extraction to obtain a set $E_{d}^{a}$ that is mostly aligned with the domain knowledge. \textbf{2) Schema-constraint Extraction}. In the vertical professional domains, the data structure between multiple documents in each data source such as drug instructions, physical examination reports, government affairs, online order data, structured data tables, etc. has strong consistency, and is more suitable for information extraction with schema-constraint, structured Extraction also makes it easier to do knowledge management and quality improvement. For detailed information about knowledge construction based on Schema-constraint, please refer to the SPG\footnote{Official site of SPG: https://spg.openkg.cn/en-US} and OneKE\cite{gui2024iepile}. This article will not introduce it in detail. It is worth noting that, as shown in figure \ref{fig:llmspg}, for the same entity type, such as \textbf{Person}, we can pre-define properties and relations such as \textit{{name, gender, placeOfBirth, (Person, hasFather, Person), (Person, hasFriend, Person)}}, and can also extract tripples directly such as \textit{(Jay Chou, spgClass, Person), (Jay Chou, constellation, Capricorn), (Jay Chou, record company, Universal Music Group)} through openIE. \textbf{3) Pre-defined Knowledge Structures By Document Type}. Professional documents such as drug instructions, government affairs documents, and legal definitions generally have a relatively standardized document structure. Each type of document can be defined as an entity type, and different paragraphs are different properties of the entity. Taking government affairs as an example, we can pre-define the GovernmentAffair EntityType and properites such as \textit{administrative divisions, service procedures, required materials, service locations, and target groups}. The divided chunks are the values of different properties. If the user asks \textit{"What materials are needed to apply for housing provident fund in Xihu District?"}, you can directly take out the chunk corresponding to property \textit{required materials} to answer the question, avoiding the possible hallucinations caused by LLM re-generation.

\subsubsection{Mutual indexing between text chunk vectors and knowledge structures}

KAG's mutual-indexing is a knowledge management and storage mechanism that conforms to the LLMFriSPG semantic representation. As is described in section 2.1, it includes four core data structures:  \textbf{1) Shared Schema}s are coarse-grained-types pre-defined as SPG Classes at project level, it includes \textit{EntityType}s, \textit{ConceptType}s, and \textit{EventType}s, they serve as a high-level categorization such as \textit{Person, Organization, GEOLocation, Date, Creature, Work, Event}. \textbf{2) Instance Graph} include all event and entity instances in $KG_{cs}$ and $KG_{fr}$. that is, instances constructed through openIE with schema-free or structured extraction with schema-constraint are both stored as instances in KG storage. \textbf{3) Text Chunks} are special entity node that conforms to the definition of the \textit{Chunk EntityType}. \textbf{4) Concept Graph} is the core component for knowledge alignment. it consists of a series of concepts and concept relations, concept nodes are also fine-grained-types of instances. Through relation prediction, instance nodes can be linked to concept nodes to obtain their fine-grained semantic types.
, and two storage structures: \textbf{1) KG Store}. Store KG data structures in LPG databases, such as TuGraph, Neo4J. \textbf{2) Vector Store}. Store text and vectors in a vector storage engine, such as ElasticSearch, Milvus, or the vector storage embedded in the LPG engine.

\subsection{Logical Form Solver}

\begin{figure*}[htbp]
    \centering
    \includegraphics[width=0.9\linewidth]{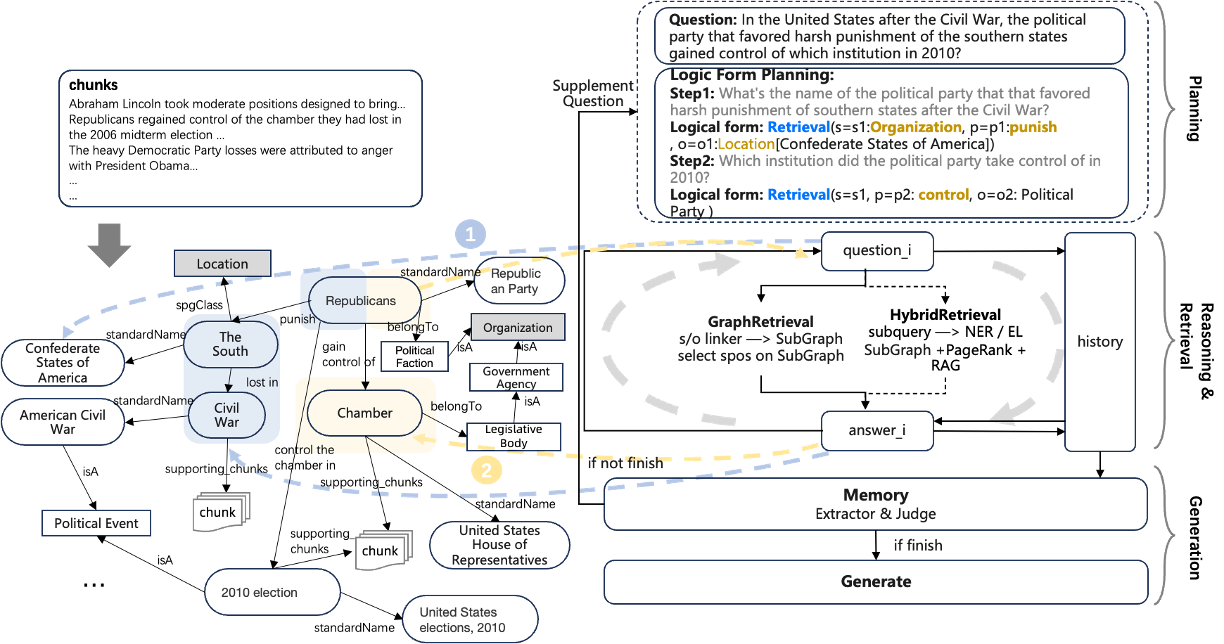}
    \caption{An Example of logical form execution. In this figure, the construction process of KG on the left is shown in Figure \ref{fig:kag-builder-example}, and the overall reasoning and iteration process is on the right. First, a logical form decomposition is performed based on the user's overall question, and then logical-form-guided reasoning is used for retrieval and reasoning. Finally, Generation determines whether the user's question is satisfied. If not, a new question is supplied to enter a new logical form decomposition and reasoning process. If it is determined to be satisfied, Generation directly outputs the answer.}
    \label{fig:example_of_logic_form}
\end{figure*}

In the process of solving complex problems, three key steps are involved: \textit{planning, reasoning} and \textit{retrieval}. Disassembling question is a planning process to determine the next problem to be tackled. Reasoning includes retrieving information based on the disassembled question, inferring the answer to the question according to the retrieved results, or re-disassembling the sub-question when the retrieved content cannot answer the question. Retrieval is to find the content that can be used as reference for the original question or the disassembled sub-question. 
\begin{algorithm}[htbp]
\caption{Logical Form Solver}
\begin{algorithmic}[1]
\State $memory \gets []$
\State $query_{cur} \gets query$ 
\For{$round \in (0,n)$}
    \State $lf_{list} \gets$ LFPlanner($query_{cur}$)
    \State $history$ $\gets$ $[]$
    \For{$lf \in lf_{list}$}
        \State $lf_{subquery}, lf_{func} \gets lf$
        \State $retrievals_{sub}$, $answer_{sub}$ $\gets$ Reasoner($lf_{subquery}$, $lf_{func}$)
        \State $history.append([lf_{subquery}, retrievals_{sub}, answer_{sub}])$
    \EndFor
    \State $memory$ $\gets$ Memory($query$, $history$)
    \If{not Judge($query$, $memory$)} 
        \State $query_{cur}$ $\gets$ SupplyQuery($query$, $memory$)
    \EndIf
\EndFor
\State $answer \gets$ Generator($query$, $memory$)
\State \Return $answer$
\label{alg:logic_form_executor}
\end{algorithmic}
\end{algorithm}
Since interactions between different modules in traditional RAG are based on vector representations of natural language, inaccuracies often arise. Inspired by the logical forms commonly used in KGQA, we designed an executable language with reasoning and retrieval capabilities. This language breaks down a question into multiple logical expressions, each of which may include functions for retrieval or logical operations. The mutual indexing described in Section 2.2 makes this process possible. Meanwhile, we designed a multi-turn solving mechanism based on reflection and global memory, inspired by ReSP\cite{jiang2024retrieve}. The KAG solving process, as referenced in Figure \ref{fig:example_of_logic_form} and Algorithm \ref{alg:logic_form_executor}, first decomposes the current question $query_{cur}$ into a list of subquestions $lf_{list}$ represented in logical form, and performs hybrid reasoning to solve them. If an exact answer can be obtained through multi-hop reasoning over structured knowledge, it returns the answer directly. Otherwise, it reflects on the solution results: storing the answers and retrieval results corresponding to $lf_{list}$ in global memory and determining whether the question is resolved. If not, it generates supplementary questions and proceeds to the next iteration. Section 2.3.1, 2.3.2 and 2.3.3 introduce logical form function for planning, logical form for reasoning and logical form for retrieval respectively. In general, the proposed logical form language has the following three advantages:
\begin{itemize}
    \item The use of symbolic language enhances the rigor and interpretability of problem decomposition and reasoning.
    \item Make full use of LLMFriSPG hierarchical representation to retrieve facts and texts knowledge guided by the symbolic graph structure
    \item Integrate the problem decomposition and retrieval processes to reduce the system complexity.
\end{itemize}   

Table \ref{tab:case-for-lf-reasoning} illustrates a multi-round scenario consistent with pseudocode \ref{alg:logic_form_executor}. Although first round the exact number of \textit{plague occurrences} couldn't be determined, but we can extracted information indicates: \textit{"Venice, the birthplace of Antonio Vivaldi, experienced the devastating Black Death, also known as the Great Plague. This pandemic caused by Yersinia pestis led to 75 to 200 million deaths in Eurasia, peaking in Europe from 1347 to 1351. The plague brought significant upheavals in Europe. Although specific occurrence records in Venice aren't detailed, it's clear the city was impacted during the mid-14th century."}. As is shown in Table \ref{tab:case-for-lf-reasoning},After two iterations, the answer determined is: \textit{\textbf{22 times}}.

\subsubsection{Logical Form Planning}
Logical Functions are defined as Table \ref{tab: logical-form-Function}, with each function representing an execution action. Complex problems are decomposed by planning a combination of these expressions, enabling reasoning about intricate issues.
\renewcommand\arraystretch{1.1}
\begin{table*}[htbp]
    \centering
    \small
\begin{tabular}{c|l}
	\toprule
	Function Name & Function Declaration\\
	\midrule
	 Retrieval & \begin{tabular}{p{10cm}} 
		$\textcolor{blue}{Retrieval}(s=s_{i}:type[name], p=p_{i}:edge, o=o_{i}:type[name],$ \\
		$s.prop=value, p.prop=value, o.prop=value$) 
	\end{tabular} \\  \hline
      Sort & \begin{tabular}{p{10cm}} 
		$\textcolor{blue}{Sort}(A, direction=min|max, limit=n)$
	\end{tabular} \\  \hline
	 Math & \begin{tabular}{p{10cm}} 
		$math_{i}=\textcolor{blue}{Math}(expr)$,  \\
		expr is in LaTeX syntax and can be used to perform operations on sets. \\
            e.g. count: $\|A\|$, sum: $\sum{A}$
	\end{tabular} \\  \hline
      Deduce & \begin{tabular}{p{10cm}} 
		$\textcolor{blue}{Deduce}(left=A, right=B, op=entailment|greater|less|equal)$
	\end{tabular} \\  \hline
      Output & \begin{tabular}{p{10cm}} 
		$\textcolor{blue}{Output}(A,B,...)$
	\end{tabular} \\
	\bottomrule
\end{tabular}
\caption{Functions of logical form. }
\label{tab: logical-form-Function}
\end{table*}

\textbf{Retrieval}. According to the the knowledge or information retrieved from SPO, \textit{s, p, o} should not repeatedly appear multiple times in the same expression. Constraints can be applied to the \textit{s, p, o} for querying. For multi-hop queries, multiple retrievals are required. When the current variable refers to a previously mentioned variable, the variable name must be consistent with the referenced variable name, and only the variable name needs to be provided. The knowledge type and name are only specified during the first reference. 

\textbf{Sort}. Sort the retrieved results. $A$ is the variable name for the retrieved subject-predicate-object(SPO) ($s_{i}$, $o_{i}$, or $s.prop$, $p.prop$, $o.prop$). $direction$ specifies the sorting direction, where $direction=min$ means sorting in ascending order and $direction=max$ means sorting in descending order. $limit=n$ indicates outputting the topN results.

\textbf{Math}. Perform mathematical calculations. $expr$ is in LaTeX syntax and can be used to perform calculations on the retrieved results (sets) or constants. $math_{i}$ represents the result of the calculation and can be used as a variable name for reference in subsequent actions.

\textbf{Deduce}. Deduce the retrieval or calculation results to answer the question. $A,B$ can be the variable names from the retrieved SPO or constants. The operator $op=entailment|greater|less|equal$ represents $A$ entails $B$, $A$ is greater than $B$, $A$ is less than $B$, and $A$ is equal to $B$, respectively.

\subsubsection{Logical Form for Reasoning}
When the query statement represented by natural language is applied to the search, the logic is often fuzzy, such as \textit{"find a picture containing vegetables or fruits"} and \textit{"find a picture containing vegetables and fruits"}. Whether text search or vector search is used, the similarity between the two queries is very high, but the corresponding answers are quite different. The same is true for problems involving logical reasoning processes such as \textit{and} or \textit{not}, and \textit{intersection} differences. To this end, we use logical form to express the question, so that it can express explicit semantic relations.
Similar to IRCOT, we decompose complex original problem and plan out various execution actions such as multi-step retrieval, numerical reasoning, logical reasoning, and semantic deduce. Each sub-problem is expressed using logical form functions, and dependencies between sub-questions are established through variable references. The inference resolution process for each sub-question is illustrated as Algorithm \ref{alg:logic_form_reasoner}. In this process, the  \textbf{GraphRetrieval} module performs KG structure retrieval according to the logical form clause to obtain structured graph results. Another key module, \textbf{HybridRetrieval}, combining natural language expressed sub-problems and logical functions for comprehensive retrieval of documents and sub-graph information. To understand how logical functions can be utilized to reason about complex problems, refer to the following examples as Table \ref{tab:case-for-reasoning}.

\textbf{Output}. Directly output $A, B, ...$ as the answers. Both $A$ and $B$ are variable names that reference the previously retrieved or calculated
\begin{algorithm*}[htbp]
\caption{Logical Form Reasoner}
\begin{algorithmic}[2]
\Require Each sub-query resulting from the decomposition of a question based on the logical form, along with their respective logical function, are denoted as $lf_{subquery}$ and $lf_{func}$
\Ensure The retrievals and answer of each sub-query, are denoted as $retri_{sub}$ and $answer_{sub}$
\State $retri_{kg} \gets$ GraphRetrieval($lf_{subquery}$, $lf_{func}$)
\If{$retri_{kg} \neq None$ and $retri_{kg} > threshold$}
    \State $retri_{sub} \gets retri_{kg}$
\Else{}
    \State $retri_{doc} \gets$ HybridRetrieval($lf_{subquery}$, $retri_{kg}$)
    \State $retri_{sub} \gets retri_{kg}, retri_{doc}$  
\EndIf
\State $answer_{sub} \gets$ Generator($lf_{subquery}$, $retri_{sub}$)
\State \Return $retri_{sub}, answer_{sub}$
\label{alg:logic_form_reasoner}
\end{algorithmic}
\end{algorithm*}

\subsubsection{Logical Form for Retrieval}
In naive RAG, retrieval is achieved by calculating the similarity (e.g. cosine similarity) between the embeddings of the question and document chunks, where the semantic representation capability of embedding models plays a key role. This mainly includes a sparse encoder (BM25) and a dense retriever (BERT architecture pre-training language models). 
Sparse and dense embedding approaches capture different relevance features and can benefit from each other by leveraging complementary relevance information.

The existing method of combining the two is generally to combine the scores of the two search methods in an ensemble, but in practice different search methods may be suitable for different questions, especially in questions requiring multi-hop reasoning. When query involves proper \textit{nouns, people, places, times, numbers, and coordinates}, the representation ability of the pre-trained presentation model is limited, and more accurate text indexes are needed. For queries that are closer to the expression of a paragraph of text, such as scenes, behaviors, and abstract concepts, the two may be coupled in some questions. 

In the design of logical form, it is feasible to effectively combine two retrieval methods. When keyword information is needed as explicit filtering criteria, conditions for selection can be specified within the retrieval function to achieve structured retrieval. 

For example, for the query \textit{"What documents are required to apply for a disability certificate at West Lake, Hangzhou?"}, the retrieval function could be represented as: \textit{"\textcolor{blue}{Retrieval}(s=s1:\textcolor{brown}{Event}[applying for a disability certificate], p=p1:\textcolor{brown}{support\_chunks}, o=o1:\textcolor{brown}{Chunk}, s.location=West Lake, Hangzhou)"}. This approach leverages the establishment of different indices (sparse or dense) to facilitate precise searches or fuzzy searches as needed.

Furthermore, when structured knowledge in the form of SPO cannot be retrieved using logical functions, alternative approaches can be employed. These include semi-structured retrieval, which involves using logical functions to search through chunks of information, and unstructured retrieval. The latter encompasses methods such as Retrieval-Augmented Generation (RAG), where sub-problems expressed in natural language are used to retrieve relevant chunks of text. This highlights the adaptability of the system to leverage different retrieval strategies based on the availability and nature of the information.

\subsection{Knowledge Alignment}
Constructing KG index through information-extraction and retrieving based on vector-similarity has three significant defects in knowledge alignment:
\begin{itemize}
    \item \textbf{Misaligned semantic relations between knowledge}. Specific semantic relations, such as \textit{contains, causes} and \textit{isA}, are often required between the correct answer and the query, while the similarity relied upon in the retrieval process is a weak semantic measure that lacks properties and direction, which may lead to imprecise retrieval of content.
    \item \textbf{Misaligned knowledge granularity}. The problems of knowledge granularity difference, noise, and irrelevance brought by openIE pose great challenges to knowledge management. Due to the diversity of language expressions, there are numerous synonymous or similar nodes, resulting in low connectivity between knowledge elements, making the retrieval recall incomplete.
    \item \textbf{Misaligned with the domain knowledge structure}. There is a lack of organized, systematic knowledge within specific domains. Knowledge that should be interrelated appears in a fragmented state, leading to a lack of professionalism in the retrieved content.
\end{itemize}
To solve these problems, we propose a solution that leverages concept graphs to enhance offline indexing and online retrieval through semantic reasoning. This involves tasks such as \textit{knowledge instance standardization, instance-to-concept linking, semantic relation completion, and domain knowledge injection}. As described in section 2.2.2, we added descriptive text information to each instance, concept or relation in the extraction phase to enhance its interpretability and contextual relevance. Meanwhile, as described in section 2.2.3, KAG supports the injection of domain concepts and terminology knowledge to reduce the noise problem caused by the mismatch of knowledge granularity in vertical domains. The goal of concept reasoning is to make full use of vector retrieval and concept reasoning to complete concept relations based on the aforementioned knowledge structure to enhance the accuracy and connectivity of the domain KG. Refer to the definition of SPG concept semantics\footnote{Semantic Classification of Concept: https://openspg.yuque.com/ndx6g9/ps5q6b/fe5p4nh1zhk6p1d8}, as is shown in Table \ref{tab:sem-rel-def}, we have summarized six semantic relations commonly required for retrieval and reasoning. Additional semantic relations can be added based on the specific requirements of the actual scenario. 

\renewcommand\arraystretch{1.1}
\begin{table*}[htbp]
    \small
    \centering
    \setlength\aboverulesep{0pt}\setlength\belowrulesep{0pt}
    \begin{tabular}{l|l|l} 
       \toprule
       Formal Expression & Description & Example \\
       \cline{1-3} 
       \midrule
       \textit{<var1, synonym, var2>}  & \begin{tabular}[c]{@{}l@{}}A \textit{synonym} relation means that a word or phrase\\ \textit{var2} that has the same or nearly the same meaning \\as another word or phrase  \textit{var1} in the same language \\and given context.\end{tabular}& \textit{Fast} is a synonym of \textit{quick}. \\
       \midrule
       \textit{<var1, isA, var2>} & \begin{tabular}[c]{@{}l@{}}An \textit{isA} relation means that a hypernym \textit{var2} that is \\more generic or abstract than a given word or phrase \\\textit{var1} and encompasses a broader category that the \\given word belongs to. \end{tabular}  &  \textit{Car} isA \textit{Vehicle}. \\
       \midrule
       \textit{<var1, isPartOf, var2>} &  \begin{tabular}[c]{@{}l@{}}An \textit{isPartOf} relation means that something \textit{var1} is a \\component or constituent of something \textit{var2} larger.\\ This relation shows that an item is a part of a \\ bigger whole. \end{tabular} &  \textit{Wheel} isPartOf \textit{car}. \\
       \midrule
       \textit{<var1, contains, var2>} &  \begin{tabular}[c]{@{}l@{}}A \textit{contains} relation means that something \textit{var1} \\includes or holds \textit{var2}, something else within it. \\This indicates that one item has the other as a subset\\ or component. \end{tabular} &  \textit{Library} contains \textit{books}. \\
       \midrule
       \textit{<var1, belongTo, var2>} &  \begin{tabular}[c]{@{}l@{}}An \textit{belongTo} relation means that something \textit{var1} is an \\instance of concept \textit{var2}. \end{tabular} &  \textit{Chamber} belongTo \textit{Legislative Body}. \\
       \midrule
       \textit{<var1, causes, var2>} &  \begin{tabular}[c]{@{}l@{}}A \textit{causes} relation means that one event or action \textit{var1} \\brings about another \textit{var2}. This indicates a causal \\relation where one thing directly results in the \\occurrence of another. \end{tabular} & \textit{Fire} causes \textit{smoke}. \\
       \bottomrule
    \end{tabular}
    \caption{Commonly used semantic relations. }
    \label{tab:sem-rel-def}
\end{table*}

\subsubsection{Enhance Indexing}
The process of enhancing indexing through semantic reasoning, as shown in Figure \ref{fig:kag-builder-example} , specifically implemented as predicting semantic relations or related knowledge elements among index items using LLM, encompassing four strategies:
\begin{itemize}
    \item \textit{Disambiguation and fusion of knowledge instances}. Taking entity instance $e_{cur}$ as an example, first, the one-hop relations and description information of $e_{cur}$ are used to predict \textit{synonymous} relations to obtain the synonym instance set $E_{syn}$ of $e_{cur}$. Then, the fused target entity $e_{tar}$ is determined from $E_{syn}$. Finally, the entity fusion rules are used to copy the properties and relations of the remaining instances in $E_{syn}$ to $e_{tar}$, and the names of these instances are added to the \textit{synonyms} of $e_{tar}$, the remaining instances will also be deleted immediately.
    \item \textit{Predict relations between instances and concepts}. For each knowledge instance (such as event, entity), predict its corresponding concept and add the derived triple $<e_i,\ belongTo,\ c_j>$ to the knowledge index. As is shown in Figure \ref{fig:kag-builder-example}, <\textit{Chamber, belongTo, Legislative Body}> means that the Chamber belongs to Legislative Body in classification.
    \item \textit{Complete concepts and relations between concepts}. During the extraction process, we use concept reasoning to complete all \textit{hypernym} and \textit{isA relation}s between semanticType and spgClass. As is shown in Figure \ref{fig:kag-builder-example} and Table \ref{tab:sem-rel-def}, we can obtain the semanticType of \textit{Chamber} is \textit{Legislative Body}, and its \textit{spgClass} is \textit{Organization} in the extraction phase. Through semantic completion, we can get \textit{<Legislative Body, isA, Government Agency>, <Government Agency, isA, Organization>}. Through semantic completion, the triple information of $KG_{fr}$ space is more complete and the connectivity of nodes is stronger.
\end{itemize}

\subsubsection{Enhance Retrieval}
In the retrieval phase, we utilize semantic relation reasoning to search the KG index based on the phrases and types in the logical form. For the types, mentions or relations in the logical form, we employ the method of combining semantic relation reasoning with similarity retrieval to replace the traditional similarity retrieval method. This retrieval method makes the retrieval path professional and logical, so as to obtain the correct answer. First, the hybrid reasoning performs precise type matching and entity linking. If the type matching fails, then, semantic reasoning is performed. As shown in Figure \ref{fig:example_of_logic_form}, if the type \textit{Political Party} fails to match, semantic reasoning is used to predict that \textit{Political Party} contains \textit{Political Faction}, and reasoning or path calculation is performed starting from \textit{Political Faction}. \newline \newline
Take another example. If the user query $q_1$ is \textit{"Which public places can cataract patients go for leisure?"} and the document content $d_2$ is \textit{"The museum is equipped with facilities to provide barrier-free visiting experience services such as touch, voice interpretation, and fully automatic guided tours for the visually impaired."}, It is almost impossible to retrieve $d_2$ based on the vector similarity with $q_1$. However, it is easier to retrieve $d_2$ through the semantic relation of \textit{<cataract patient, isA, visually impaired>}.

\subsection{KAG-Model}
KAG includes two main computational processes: \textit{offline index building} and \textit{online query and answer generation}. In the era of \textit{small language models}, these two tasks were typically handled by two separate pipelines, each containing multiple task-specific NLP models. This results in high complexity for the application system, increased setup costs, and inevitable cascading losses due to error propagation between modules. In contrast, large language models, as a \textit{capability complex}, can potentially integrate these pipelines into a unified, simultaneous end-to-end reasoning process.

\begin{figure}[htbp]
    \centering
    \includegraphics[width=0.9\linewidth]{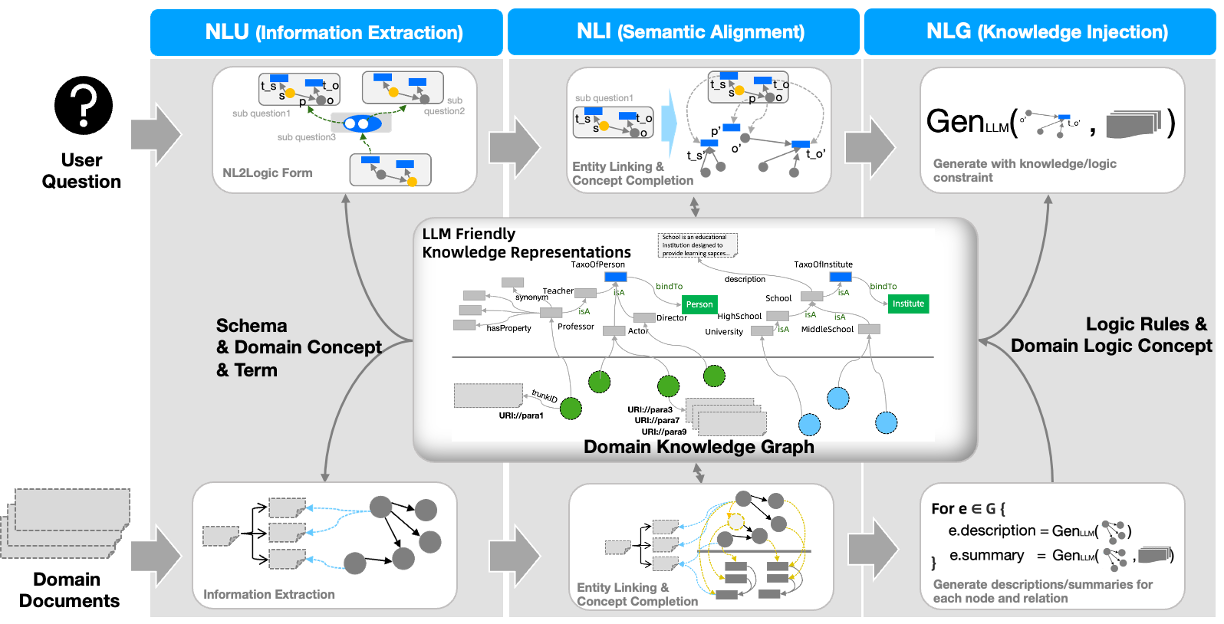}
    \caption{The model capabilities required for KAG.}
    \label{fig:Synergy-nlp-kg}
\end{figure}

As shown in Figure \ref{fig:Synergy-nlp-kg}, the processes of indexing and QA each consist of similar steps. Both of the two pipelines can be abstracted as \textit{classify, mention detection, mention relation detection, semantic alignment, embedding}, and \textit{chunk, instance, or query-focused summary}. Among these, \textit{classify, mention detection}, and \textit{mention relation detection} can be categorized as NLU, while \textit{semantic alignment} and \textit{embedding} can be grouped under NLI. Finally, the \textit{chunk, instance or query-focused summary} can be classified under NLG. Thus, we can conclude that the three fundamental capabilities of natural language processing that a RAG system relies on are NLU, NLI, and NLG.

We focused on exploring methods to optimize these three capabilities, which are introduced in subsections 2.5.1, 2.5.2, and 2.5.3 respectively. Additionally, to reduce the cascade loss caused by linking models into a pipeline, we further explored methods to integrate multiple inference processes into a single inference. Subsection 2.5.4 will discuss how to equip the model with retrieval capabilities to achieve better performance and efficiency through one-pass inference.

\subsubsection{Natural Language Understanding}
NLU is one of the most common foundational tasks in natural language processing, including \textit{text classification, named entity recognition, relation Extraction, subject and object extraction, trigger detection, event argument extraction, event extraction}, and \textit{machine reading comprehension}. We have collected over 30 public datasets to enhance understanding capabilities. Experiments found that simply transforming the original datasets into instruction datasets can achieve comparable results to specialized models on trained tasks, but this approach does not improve the model's NLU capabilities on unseen domains. Therefore, we conducted large-scale instruction reconstruction, designing various instruction synthesis strategies to create an NLU instruction dataset with over 20,000 diverse instructions. By utilizing this dataset for supervised fine-tuning on a given base model, the model has demonstrated enhanced NLU capabilities in downstream tasks. The instruction reconstruction strategy mainly consists of the following three types.
\begin{itemize}
    \item \textbf{Label bucketing:} \cite{gui2024iepile}This strategy focuses on label-guided tasks, where the aim is to extract text based on labels or map text to specified labels, including \textit{classification, NER, RE}, and \textit{EE}. When labels in a dataset collectively co-occur in the training set, the model may learn this pattern and overfit to the dataset, failing to independently understand the meaning of each label. Therefore, during the instruction synthesis process, we adopt a polling strategy that designates only one label from each training sample as part of a bucket. Additionally, since some labels have similar semantics and can be confused, we group easily confused labels into a single bucket, allowing the model to learn the semantic differences between the two labels more effectively.
    \item \textbf{Flexible and Diverse Input and Output Formats:} The LLM employs an instruction-following approach for inference, and a highly consistent input-output format may cause the model to overfit to specific tasks, resulting in a lack of generalization for unseen formats. Therefore, we have flexibly processed the input and output formats. The output is handled as five different formatting instructions, as well as two types of natural language instructions. Additionally, the output format can dynamically be specified as markdown, JSON, natural language, or any format indicated in the examples.
    \item \textbf{Instructoin with Task Guildline:} Traditional NLP training often employs a "sea of questions" approach, incorporating a wide variety of data in the training set. This allows the model to understand task requirements during the learning process, such as whether to include job titles when extracting personal names. For the training of LLMs, we aim for the model to perform tasks like a professional annotator by comprehending the task description. Therefore, for the collected NLU tasks, we summarize the task descriptions using a process of self-reflection within the LLM. This creates training data that includes task descriptions within the instructions. Additionally, to enhance task diversity, we implement heuristic strategies to rephrase the task descriptions and answers. This enables the model to understand the differences between task descriptions more accurately and to complete tasks according to the instructions.
\end{itemize}
We fine-tuned six foundational models: qwen2, llama2, baichuan2, llama3, mistral, phi3, and used six understanding benchmarks recorded on OpenCompass for performance validation. The table \ref{Tab.llm_nlu} shows that the KAG-Model has a significant improvement in NLU tasks.

\renewcommand\arraystretch{1.1}
\begin{table}[htbp]
\centering
\small
\setlength\aboverulesep{0pt}\setlength\belowrulesep{0pt}
\begin{tabular*}{\textwidth}{@{\extracolsep{\fill}} l|ccccccc}
\toprule
     Models & C3  & WSC & XSum  & Lambda  & Lcsts    & Race  & Average   \\ 
     \midrule
GPT4                  & 95.10          & 74.00          & 20.10          & 65.50          & 12.30          & 92.35        & 59.89  \\
\midrule
Qwen2                 & 92.27          & 66.35          & 18.68          & 62.39          & 13.07          & \textbf{88.37}    & 56.86      \\
KAG$_{\rm Qwen2}$     & \textbf{92.88} & \textbf{70.19} & \textbf{31.33} & \textbf{66.16} & \textbf{18.53} & 88.17    & \textbf{61.21}  \\ 
\midrule
Llama2                & 81.70           & 50.96          & 23.29          & 63.26          & 15.99          & 55.64         &48.47  \\ 
KAG$_{\rm Llama2}$    & \textbf{82.36} & \textbf{63.46} & \textbf{24.51} & \textbf{65.22} & \textbf{17.51} & \textbf{68.48}  &\textbf{53.59} \\
\midrule
Baichuan2             & \textbf{84.44} & 66.35          & 20.81          & 62.43          & 16.54          & 76.85        &54.57  \\ 
KAG$_{\rm Baichuan2}$ & 84.11          & \textbf{66.35} & \textbf{21.51} & \textbf{62.64} & \textbf{17.27} & \textbf{77.18}   & \textbf{54.84} \\
\midrule
Llama3                & \textbf{86.63} & \textbf{65.38} & 25.84          & 36.72          & 0.09           & \textbf{83.76}  & 49.74 \\ 
KAG$_{\rm Llama3}$    & 83.40          & 62.50          & \textbf{26.72} & \textbf{54.07} & \textbf{18.45} & 81.16     & \textbf{54.38}      \\
\midrule
Mistral               & \textbf{67.29} & 30.77          & 21.16          & 59.98          & 0.78      & 73.46    & 42.24      \\ 
KAG$_{\rm Mistral}$   & 47.29          & \textbf{39.42} & \textbf{21.54} & \textbf{69.09} & \textbf{17.14} & 72.42       & \textbf{44.48}   \\ 
\midrule
Phi3                  & 68.60          & \textbf{42.31} & \textbf{0.60}  & \textbf{71.74} & 3.47           & 73.18     & 43.32     \\ 
KAG$_{\rm Phi3}$      & \textbf{85.21} & 25.94          & 0.36           & 71.24          & \textbf{15.49} & \textbf{74.00}     & \textbf{45.37} \\  
 \bottomrule
 \end{tabular*}
\caption{Enhancement of natural language understanding capabilities in different LLMs by KAG. The experimental results are based on the open-compass framework and tested using  the ``gen'' mode. The evaluation metrics for C3, WSC, Lambda, and Race are ACC. XSum and Lcsts are measured using ROUGE-1. Race includes Race-middle and Race-high, and their average is taken.} 
\label{Tab.llm_nlu} 
\end{table}

\subsubsection{Natural Language Inference}

The NLI task is used to infer the semantic relations between given phrases. Typical NLI tasks include \textit{entity linking, entity disambiguation, taxonomy expansion, hypernym discovery}, and \textit{text entailment}. In the context of knowledge base Q\&A, due to the diversity and ambiguity of natural language expressions, as well as the subtle and different types of semantic connections between phrases, it often requires further alignment or retrieval of related information through NLI tasks based on NLU.
As described in section 2.4, we categorize the key semantic relation in knowledge base applications into six types. Among these, relations such as \textit{isA, isPartOf} and \textit{contains} exhibit directional and distance-based partial order relations. During the reasoning process, it is crucial to accurately determine these semantic relations to advance towards the target answer. In traditional approaches, separate training of representation pre-training models and KG completion(KGC) models is often employed to reason about semantic relations. However, these KGC models tend to focus on learning graph structures and do not fully utilize the essential textual semantic information for semantic graph reasoning. LLMs possess richer intrinsic knowledge, and can leverage both semantic and structural information to achieve more precise reasoning outcomes. To this end, we have collected a high-quality conceptual knowledge base and ontologies from various domains, creating a conceptual knowledge set that includes 8,000 concepts and their semantic relations. Based on this knowledge set, we constructed a training dataset that includes six different types of conceptual reasoning instructions to enhance the semantic reasoning capabilities of a given base model, thereby providing semantic reasoning support for KAG.

Semantic reasoning is one of the core ability required in KAG process, we use NLI tasks and general reasoning Q\&A tasks to evaluate the ability of our model, the results are as shown in Table \ref{Tab.model_nli} and Table \ref{Tab:model-nli-task}. The evaluation results indicates that our KAG-Model demonstrates a significant improvement in tasks related with semantic reasoning: First, Table \ref{Tab:model-nli-task} shows that on the Hypernym Discovery task(which is consistent in form with the reasoning required in semantic enhanced indexing and retrieval.), our fine-tuned KAG-llama model outperforms Llama3 and ChatGPT-3.5 significantly. In addition, the better performance of our model on CMNLI, OCNLI and SIQA compared with Llama3 in Table \ref{Tab.model_nli} shows that our model has good capabilities in general logical reasoning. 

\renewcommand\arraystretch{1.1}
\begin{table}[htbp]
\centering
\small
\setlength\aboverulesep{0pt}\setlength\belowrulesep{0pt}
\begin{tabular*}{\textwidth}{@{\extracolsep{\fill}} l|ccc}
\toprule
     Models & CMNLI  & OCNLI & SIQA   \\ 
     \midrule
    Llama3       & 35.14          & 32.1          & 44.27  \\
    KAG-Llama3   & 49.52          & 44.31         & 65.81  \\

 \bottomrule
 \end{tabular*}
\caption{Enhancement of natural language Inference capabilities in different LLMs by KAG. The evaluation metrics for CMNLI, OCNLI, SIQA are measured with accuracy. } 
\label{Tab.model_nli} 
\end{table}

\renewcommand\arraystretch{1.1}
\begin{table*}[htbp]
    \small
    \centering
    \setlength\aboverulesep{0pt}\setlength\belowrulesep{0pt}
    \begin{tabular*}{\textwidth}{@{\extracolsep{\fill}} l|ccc}
       \toprule
        & 1A.English & 2A.Medical & 2B.Music \\
       \midrule
       ChatGPT-3.5 & {\ul 30.04} & {\ul 26.12} & {\ul 28.47}  \\
       Llama3-8B & 23.47 & 24.26  & 18.73 \\
        \midrule
       KAG-Llama3 & \textbf{38.26} & \textbf{55.14} & \textbf{30.16} \\
       
       \bottomrule
    \end{tabular*}
    \caption{Hypernym Discovery performance comparison on SemEval2018-Task9 dataset, measured in MRR.}
    \label{Tab:model-nli-task}
\end{table*}

\subsubsection{Natural Language Generation}
Models that have not undergone domain adaptation training often exhibit significant differences from the target text in domain logic and writing style. Moreover, acquiring sufficient amounts of annotated data in specialized domains frequently poses a challenge. Therefore, we have established two efficient fine-tuning methods for specific domain scenarios, allowing the generation process to better align with scene expectations: namely, K-Lora and AKGF.

\textbf{Pre-learning with K-LoRA}. First of all, we think that using knowledge to generate answers is the reverse process of extracting knowledge from text. Therefore, by inverting the previously described extraction process, we can create a 'triples-to-text' generation task. With extensive fine-tuning on a multitude of instances, the model can be trained to recognize the information format infused by the KG. Additionally, as the target text is domain-specific, the model can acquire the unique linguistic style of that domain. Furthermore, considering efficiency, we continue to utilize LoRA-based SFT. We refer to the LoRA obtained in this step as K-LoRA.

\textbf{Alignment with KG Feedback}. The model may still exhibit hallucinations in its responses due to issues such as overfitting. Inspired by the RLHF(Reinforcement Learning with Human Feedback) approach\cite{ouyang2022training,ziegler2019fine}, we hope that the KG can serve as an automated evaluator, providing feedback on knowledge correctness of the current response, thereby guiding the model towards further optimization. First, we generate a variety of responses for each query by employing diverse input formats or random seeds. Subsequently, we incorporate the KG to score and rank these responses. The scoring process compare generated answer with knowledge in KG to ascertain their correctness. The reward is determined by the number of correctly matched knowledge triples. The formula for calculating the reward is represented by Formula 1. 
    
    \begin{center}
    $reward$ = $\log(rspo + \alpha \times re) \ \  (1)$   
    \end{center}
    where $\alpha$ is a hyperparameter, \textit{rspo} represents the number of SPO matches, and \textit{re} represents the number of entity matches. 

We select two biomedical question-answering datasets, CMedQA\cite{cmedqa} and BioASQ\cite{bioasq}, for evaluating our model. CMedQA is a comprehensive dataset of Chinese medical questions and answers, while BioASQ is an English biomedical dataset. We randomly choose 1,000 instances from each for testing. For CMedQA, we employ the answer texts from the non-selected Q\&A pairs as corpora to construct a KG in a weakly supervised manner. Similarly, with BioASQ, we use all the provided reference passages as the domain-specific corpora. Experimental results, as shown in Table \ref{tab:model-nlg}, demonstrate significant enhancement in generation performance. For more details on the specific implementation process, please refer to our paper\cite{akgf}
\renewcommand\arraystretch{1.1}
\begin{table*}[htbp]
    \small
    \centering
    \setlength\aboverulesep{0pt}\setlength\belowrulesep{0pt}
    \begin{tabular*}{\textwidth}{@{\extracolsep{\fill}} l|cc|cc}
       \toprule
       \multirow{2}{*}{Model} & \multicolumn{2}{c|}{CMedQA} & \multicolumn{2}{c}{BioASQ} \\
       \cline{2-5} 
       & Rouge-L & BLEU & Rouge-L & BLEU \\
       \midrule
       ChatGPT-3.5 0-shot  & 14.20 & 1.78 & 21.14 & 5.93\\
       ChatGPT-3.5 2-shot  & 14.66 & 2.53 & 21.42 & 6.11\\
       Llama2  & 14.02 & 2.86 & 23.47 & 7.11 \\
        \midrule
       KAG$_{\rm Llama2}$ & \textbf{15.44} & \textbf{3.46} & \textbf{24.21} & \textbf{7.79} \\
       \midrule
       
       \bottomrule
    \end{tabular*}
    \caption{Performance comparison on CMedQA \& BioASQ. "CP" indicates "continual pre-trained". We consider continual pre-training as a basic method of domain knowledge infusion, on par with other retrieval-based methods. Consequently, we do not report on the outcomes of hybrid approaches.}
    \label{tab:model-nlg}
\end{table*}

\subsubsection{Onepass Inference}
Most retrieval enhanced systems operate in a series of presentation models, retrievers, and generation models, resulting in high system complexity, construction costs, and the inevitable concatenation loss caused by error transfer between modules. We introduces an efficient one-pass unified generation and retrieval (OneGen)  model to enable an arbitrary LLM to generate and retrieve in one single forward pass. Inspired by the latest success in LLM for text embedding, we expand the original vocabulary by adding special tokens (i.e. retrieval tokens), and allocate the retrieval task to retrieval tokens generated in an autoregressive manner. During training, retrieval tokens only participate in representation fine-tuning through contrastive learning, whereas other output tokens are trained using language model objectives. At inference time, we use retrieval tokens for efficient retrieving on demand.  Unlike the previous pipeline approach where at least two models are needed for retrieval and generation, OneGen unified them in one model, thus eliminating the need for a separate retriever and greatly reducing system complexity. 

As shown in experiment results in Table \ref{tab:model-onepass}, we draw the following conclusions: 
(1) OneGen demonstrates efficacy in ${R \rightarrow G}$ task, and joint training of retrieval and generation yields performance gains on the RAG task. The Self-RAG endows LLMs with self-assessment and adaptive retrieval, while OneGen adds self-retrieval. Our method outperforms the original Self-RAG across all datasets, especially achieving improvements of 3.1pt on Pub dataset and 2.8pt on ARC dataset, validating the benefits of joint training.
(2) OneGen is highly efficient in training, with instruction-finetuned LLMs showing strong retrieval capabilities with minimal additional tuning. It requires less and lower-quality retrieval data, achieving comparable performance with just 60K noisy samples and incomplete documents, without synthetic data. For more details on the specific implementation process, please refer to paper\cite{onepass}

\renewcommand\arraystretch{1.1}
\begin{table*}[htbp]
    \small
    \centering
    \setlength\aboverulesep{0pt}\setlength\belowrulesep{0pt}
    \begin{tabular*}{\textwidth}{@{\extracolsep{\fill}} ll|cccc|cccc}
    \toprule
    \multicolumn{1}{l}{} & \multicolumn{1}{l}{} & \multicolumn{4}{c}{\textbf{Generation Performance}} & \multicolumn{2}{c}{\textbf{Retrieval Performance}} \\  \cmidrule(lr){3-6} \cmidrule(lr){7-8} 
    \multicolumn{1}{l}{} & \multicolumn{1}{l}{} & \multicolumn{2}{c}{HotpotQA} & \multicolumn{2}{c}{2WikiMultiHopQA} & HotpotQA & 2WikiMultiHopQA \\  
    \cmidrule(lr){3-4} \cmidrule(lr){5-6} \cmidrule(lr){7-7} \cmidrule(lr){8-8}
    \textbf{BackBone} & \textbf{Retriever} & EM & F1 & EM & F1 & Recall@1 & Recall@1 \\  
    \midrule
    & \multicolumn{1}{c|}{Contriever} & 52.83 & 65.64 & 70.02 & \multicolumn{1}{c|}{74.35} & 73.76 & 68.75 \\
    \multirow{-2}{*}{Llama2-7B} & \multicolumn{1}{c|}{\underline{self}} & \underline{54.82} & \underline{67.93} & \underline{75.02} & \multicolumn{1}{c|}{\underline{78.86}} & \underline{75.90} & \underline{69.79} \\  \midrule
     & \multicolumn{1}{c|}{Contriever} & 53.72 & 66.46 & 70.92 & \multicolumn{1}{c|}{75.29} & 69.79 & 66.80 \\
    \multirow{-2}{*}{Llama3.1-7B} & \multicolumn{1}{c|}{\underline{self}} & \underline{55.38} & \underline{68.35} & \underline{75.88} & \multicolumn{1}{c|}{\underline{79.60}} & \underline{72.55} & \underline{68.98} \\  \midrule
     & \multicolumn{1}{c|}{Contriever} & 48.55 & \textbf{61.02} & 68.32 & \multicolumn{1}{c|}{72.66} & 72.41 & 67.70 \\
    \multirow{-2}{*}{Qwen2-1.5B} & \multicolumn{1}{c|}{\underline{self}} & \underline{48.75} & \underline{60.98} & \underline{73.84} & \multicolumn{1}{c|}{\underline{77.44}} & \underline{72.70} & \underline{69.27} \\  \midrule
     & \multicolumn{1}{c|}{Contriever} & 53.32 & 66.22 & 70.80 & \multicolumn{1}{c|}{74.86} & 74.15 & 69.01 \\
    \multirow{-2}{*}{Qwen2-7B} & \multicolumn{1}{c|}{\underline{self}} & \underline{55.12} & \underline{67.60} & \underline{76.17} & \multicolumn{1}{c|}{\underline{79.82}} & \underline{75.68} & \underline{69.96} \\ 
    \bottomrule
    \end{tabular*}
    \caption{In RAG for Multi-Hop QA settings, performance comparison across different datasets using different LLMs.}
    \label{tab:model-onepass}
\end{table*}

\section{Experiments}
\subsection{Experimental Settings}
\textbf{Datasets.}
To evaluate the effectiveness of the KAG for knowledge-intensive question-answering task, we perform experiments on 3 widely-used multi-hop QA datasets, including HotpotQA~\cite{hotpotqa}, 2WikiMultiHopQA~\cite{2wiki}, and MuSiQue~\cite{musique}. For a fair comparison, we follow IRCoT~\cite{trivedi2022interleaving} and HippoRAG~\cite{gutierrez2024hipporag} utilizing 1,000 questions from each validation set and using the retrieval corpus related to selected questions. 

\textbf{Evaluation Metric.}
When evaluating QA performance, we use two metrics: Exact Match (EM) and F1 scores. For assessing retrieval performance, we calculate the hit rates based on the Top 2/5 retrieval results, represented as Recall@2 and Recall@5.

\textbf{Comparison Methods.}
We evaluate our approach against several robust and commonly utilized retrieval RAG methods.
\textbf{NativeRAG} using ColBERTv2~\cite{ColBERTv2} as retriever and directly generates answers based on all retrieved documents~\cite{baserag}.
\textbf{HippoRAG} is a RAG framework inspired by human long-term memory that enables LLMs to continuously integrate knowledge across external documents. In this paper, we also use ColBERTv2~\cite{ColBERTv2} as its retriever~\cite{gutierrez2024hipporag}.
\textbf{IRCoT} interleaves chain-of-thought (CoT) generation and knowledge retrieval steps in order to guide the retrieval by CoT and vice-versa. This interleaving allows retrieving more relevant information for later reasoning steps.  It is a key technology for implementing multi-step retrieval in the existing RAG framework.

\subsection{Experimental Results}

\subsubsection{Overall Results}
The end-to-end Q\&A performance is shown in Table~\ref{PerformanceOfQA}. Within the RAG frameworks leveraging ChatGPT-3.5 as backbone model, HippoRAG demonstrates superior performance compared to NativeRAG. HippoRAG employs a human long-term memory strategy that facilitates the continuous integration of knowledge from external documents into LLMs, thereby significantly enhancing Q\&A capabilities. However, given the substantial economic costs associated with utilizing ChatGPT-3.5, we opted to use the DeepSeek-V2 API as a viable alternative. On average, the performance of the IRCoT + HippoRAG configuration utilizing the DeepSeek-V2 API slightly surpasses that of ChatGPT-3.5. Our constructed framework KAG shows significant performance improvement compared to IRCoT + HippoRAG, with EM increases of 11.5\%, 19.8\%, and 10.5\% on HotpotQA, 2WikiMultiHopQA, and MuSiQue respectively, and F1 improvements of 12.5\%, 19.1\%, and 12.2\%. These advancements in end-to-end performance can largely be attributed to the development of more effective indexing, knowledge alignment and hybrid solving libraries within our framework.
We evaluate the effectiveness of the single-step retriever and multi-step retriever, with the retrieval performance shown in Table~\ref{Tab.retriever_eval}. From the experimental results, it is evident that the multi-step retriever generally outperforms the single-step retriever. Analysis reveals that the content retrieved by the single-step retriever exhibits very high similarity, resulting in an inability to use the single-step retrieval outcomes to derive answers for certain data that require reasoning. The multi-step retriever alleviates this issue. Our proposed KAG framework directly utilizes the multi-step retriever and significantly enhances retrieval performance through strategies such as mutual-indexing, logical form solving, and knowledge alignment. 
\renewcommand\arraystretch{1.1}
\begin{table*}[htbp]
\centering
\small
\setlength\aboverulesep{0pt}\setlength\belowrulesep{0pt}
\begin{tabular}{l|p{1.9cm}<{\centering}|p{1.0cm}<{\centering} p{1.0cm}<{\centering}|p{1.0cm}<{\centering} p{1.0cm}<{\centering}|p{1.0cm}<{\centering} p{1.0cm}<{\centering}}
    \toprule
    \multirow{2}{*}{\textbf{Framework}}                      & \multirow{2}{*}{\textbf{Model}} & \multicolumn{2}{c|}{HotpotQA}  & \multicolumn{2}{c|}{2WikiMultiHopQA} & \multicolumn{2}{c}{MuSiQue}   \\ \cline{3-8}
     &   & EM & F1 & EM & F1 & EM & F1  \\  
    \midrule
    NativeRAG~\cite{baserag,ColBERTv2}& ChatGPT-3.5 & 43.4 & 57.7 & 33.4 & 43.3 & 15.5 & 26.4 \\
    HippoRAG~\cite{gutierrez2024hipporag,ColBERTv2}& ChatGPT-3.5 & 41.8 & 55.0 & 46.6 & 59.2 & 19.2 & 29.8 \\ \hline
    IRCoT+NativeRAG & ChatGPT-3.5 & 45.5 & 58.4 & 35.4 & 45.1 & 19.1 & 30.5 \\
    IRCoT+HippoRAG & ChatGPT-3.5 & 45.7 & 59.2 & 47.7 & 62.7 & 21.9 & 33.3 \\
    \hline
    IRCoT+HippoRAG & DeepSeek-V2 & 51.0 & 63.7 & 48.0 & 57.1 & 26.2 & 36.5 \\
    \textbf{KAG \textit{w/} $LFS_{ref_{3}}$} & DeepSeek-V2 & \underline{59.8} & \underline{74.0} & \underline{66.3} & \underline{76.1} & \underline{35.4} & \underline{48.2} \\
    \textbf{KAG \textit{w/} $LFSH_{ref_{3}}$} & DeepSeek-V2 & \textbf{62.5} & \textbf{76.2} & \textbf{67.8} & \textbf{76.2} & \textbf{36.7} & \textbf{48.7} \\
    \bottomrule
 \end{tabular}
\caption{The end-to-end generation performance of different RAG models on three multi-hop Q\&A datasets. The values in \textbf{bold} and \underline{underline} are the best and second best indicators respectively.} 
\label{PerformanceOfQA} 
\end{table*}


\renewcommand\arraystretch{1.1}
\begin{table*}[htbp]
\centering
\small
\setlength\aboverulesep{0pt}\setlength\belowrulesep{0pt}
\renewcommand{\arraystretch}{1.2} 
\begin{tabular}{p{0.4cm}|p{2.8cm}|>{\centering\arraybackslash}p{1.2cm} >{\centering\arraybackslash}p{1.2cm}|>{\centering\arraybackslash}p{1.2cm} >{\centering\arraybackslash}p{1.2cm}|>{\centering\arraybackslash}p{1.2cm} >{\centering\arraybackslash}p{1.2cm}}
\toprule
\multirow{2}{*}{} & \multirow{2}{*}{\textbf{Retriever}} & \multicolumn{2}{c|}{\textbf{HotpotQA}} & \multicolumn{2}{c|}{\textbf{2Wiki}} & \multicolumn{2}{c}{\textbf{MuSiQue}} \\ \cline{3-8}
                  &                                     & Recall@2 & Recall@5 & Recall@2 & Recall@5 & Recall@2 & Recall@5 \\ \midrule
\multirow{7}{*}{\rotatebox{90}{Single-step}} 
                  & BM25~\cite{BM25}                    & 55.4     & 72.2     & 51.8     & 61.9     & 32.3     & 41.2     \\
                  & Contriever~\cite{contriever}        & 57.2     & 75.5     & 46.6     & 57.5     & 34.8     & 46.6     \\
                  & GTR~\cite{GTR}                      & 59.4     & 73.3     & 60.2     & 67.9     & 37.4     & 49.1     \\
                  & RAPTOR~\cite{raptor}                & 58.1     & 71.2     & 46.3     & 53.8     & 35.7     & 45.3     \\
                  & Proposition~\cite{Proposition}      & 58.7     & 71.1     & 56.4     & 63.1     & 37.6     & 49.3     \\
                  & NativeRAG~\cite{baserag,ColBERTv2}  & 64.7     & 79.3     & 59.2     & 68.2     & 37.9     & 49.2     \\
                  & HippoRAG~\cite{gutierrez2024hipporag,ColBERTv2} & 60.5     & 77.7     & 70.7     & 89.1     & 40.9     & 51.9     \\ 
                  \midrule
\multirow{5}{*}{\rotatebox{90}{Multi-step}}  
                  & IRCoT + BM25                        & 65.6     & 79.0     & 61.2     & 75.6     & 34.2     & 44.7     \\
                  & IRCoT + Contriever                  & 65.9     & 81.6     & 51.6     & 63.8     & 39.1     & 52.2     \\
                  & IRCoT + NativeRAG                   & \underline{67.9} & 82.0 & 64.1 & 74.4 & 41.7 & 53.7 \\
                  & IRCoT + HippoRAG                    & 67.0 & \underline{83.0} & \textbf{75.8} & \textbf{93.9} & \underline{45.3} & \underline{57.6} \\
                  & \textbf{KAG}                          & \textbf{72.8} & \textbf{88.8} & \underline{65.4} & \underline{91.9} & \textbf{48.5} & \textbf{65.7} \\ 
                  \bottomrule
\end{tabular}
\caption{The performance of different retrieval models on three multi-hop Q\&A datasets} 
\label{Tab.retriever_eval} 
\end{table*}

\subsection{Ablation Studies}

The objective of this experiment is to deeply investigate the impact of the knowledge alignment and logic form solver on the final results. We conduct ablation studies for each module by substituting different methods and analyzing the changes in outcomes.

\subsubsection{Knowledge Graph Indexing Ablation}

In the graph indexing phase, we propose the following two substitution methods:  

\textbf{1) Mutual Indexing Method.}
As a baseline method of KAG, according to the introduction in Sections 2.1 and 2.2, we use information extraction methods (such as OpenIE) to extract phrases and triples in document chunks, and form the mutual-indexing between graph structure and text chunks according to the hierarchical representation of LLMFriSPG, and then write them into KG storage. We denote this method as \textbf{M\_Indexing}. 

\textbf{2) Knowledge Alignment Enhancement.}
This method uses knowledge alignment to enhance the KG mutual-indexing and the logical form-guided reasoning \& retrieval. According to the introduction in Section 2.4, it mainly completes tasks such as the classification of instances and concepts, the prediction of hypernyms/hyponyms of concepts, the completion of the semantic relationships between concepts, the disambiguation and fusion of entities, etc., which enhances the semantic distinction of knowledge and the connectivity between instances, laying a solid foundation for subsequent reasoning and retrieval guided by logical forms. We denote this method as \textbf{K\_Alignment}.

\subsubsection{Reasoning and Retrieval Ablation}

\textbf{Multi-round Reflection.} We adopted the multi-round reflection mechanism from ReSP\cite{jiang2024retrieve} to assess whether the Logical Form Solver has fully answered the question. If not, supplementary questions are generated for iterative solving until the information in global memory is sufficient. We analyzed the impact of the maximum iteration count $n$ on the results, denoted as $ref_{n}$. If $n=1$, it means that the reflection mechanism is not enabled. In the reasoning and retrieval phase, we design the following three substitution methods: \newline

\textbf{1) Chunks Retriever.} We define KAG's baseline retrieval strategy with reference to HippoRAG's\cite{gutierrez2024hipporag} retrieval capabilities, with the goal of recalling the top\_k chunks that support answering the current question. The Chunk score is calculated by weighting the vector similarity and the personalized pagerank score. We denote this method as $ChunkRetri$, we denote ChunkRetri with n-round reflections as $CR_{ref_{n}}$. 

\textbf{2) Logical Form Solver (Enable Graph Retrieval).} Next, we employ a Logical Form Solver for reasoning. This method uses pre-defined logical forms to parse and answer questions. First, it explores the reasoning ability of the KG structure in $KG_{cs}$ and $KG_{fr}$ spaces, focusing on accuracy and rigor in reasoning. Then, it uses supporting\_chunks in $RC$ to supplement retrieval when the previous step of reasoning has no results. We denote this method as $LFS_{ref_{n}}$. The parameter $n$ is maximum number of iteration parameter.
   
\textbf{3) Logical Form Solver (Enable Hybrid Retrieval).} In order to make full use of the mutual-indexing structure between $KG_{fr}$ and $RC$ to further explore the role of KG structure in enhancing chunk retrieval, we modify the $LFS_{ref_{n}}$ by disabling the Graph Retrieval functionality for direct reasoning. Instead, all answers are generated using the Hybrid Retrieval method. This approach enables us to evaluate the contribution of graph retrieval to the performance of reasoning. We denote this method as $LFSH_{ref_{n}}$.

Through the design of this ablation study, we aim to comprehensively and deeply understand the impact of different graph indexing and reasoning methods on the final outcomes, providing strong support for subsequent optimization and improvement.

\subsubsection{Experimental Results and Discussion}

\begin{table*}[htbp]
\centering
\footnotesize
\renewcommand{\arraystretch}{1.2}
\setlength\aboverulesep{0pt}\setlength\belowrulesep{0pt}
\begin{tabular}{l|m{2.2cm}<{\centering}|*{6}{>{\centering\arraybackslash}p{0.08\textwidth}}}
    \toprule
    \multirow{2}{*}{\textbf{Graph Index}} & \multirow{2}{*}{\textbf{Reasoning}} & \multicolumn{2}{c|}{\textbf{HotpotQA}} & \multicolumn{2}{c|}{\textbf{2Wiki}} & \multicolumn{2}{c}{\textbf{MuSiQue}} \\ 
    \cline{3-8}
     &   & EM & F1 & EM & F1 & EM & F1  \\  
    \midrule
    \multirow{1}{*}{\textbf{M\_Indexing}} & $CR_{ref_{3}}$ & 52.4 & 65.4 & 48.2 & 56.0 & 24.6 & 36.6 \\ 
    \midrule
    \multirow{3}{*}{\textbf{K\_Alignment}} 
    & $CR_{ref_{3}}$ & 54.7 & 69.5 & 62.7 & 72.5 & 29.6 & 41.1 \\ 
    & $LFS_{ref_{1}}$ & 59.1 & 73.4 & 65.2 & 74.4 & 31.3 & 43.4 \\
    & $LFS_{ref_{3}}$ & 59.8 & 74.0 & \underline{66.3} & \underline{76.1} & \underline{35.4} & \underline{48.2} \\
    & $LFSH_{ref_{1}}$ & \underline{61.5} & \underline{76.0} & 66.0 & 75.0 & 33.5 & 44.3 \\ 
    & $LFSH_{ref_{3}}$ & \textbf{62.5} & \textbf{76.2} & \textbf{67.8} & \textbf{76.2} & \textbf{36.7} & \textbf{48.7} \\ 
    \bottomrule
\end{tabular}
\caption{The end-to-end generation performance of different model methods on three multi-hop Q\&A datasets. The backbone model is DeepSeek-V2 API. As is described in Algorithm \ref{alg:logic_form_executor}, $ref_{3}$ represents a maximum of 3 rounds of reflection, and $ref_{1}$ represents a maximum of 1 round, which means that no reflection is introduced.}
\label{tab:AblationPerformanceOfQA}
\end{table*}

\begin{table*}[htbp]
\centering
\footnotesize
\renewcommand{\arraystretch}{1.2}
\setlength\aboverulesep{0pt}\setlength\belowrulesep{0pt}
\begin{tabular}{l|m{2.2cm}<{\centering}|*{6}{>{\centering\arraybackslash}p{0.08\textwidth}}}
    \toprule
    \multirow{2}{*}{\textbf{Graph Index}} & \multirow{2}{*}{\textbf{Reasoning}} & \multicolumn{2}{c|}{\textbf{HotpotQA}} & \multicolumn{2}{c|}{\textbf{2Wiki}} & \multicolumn{2}{c}{\textbf{MuSiQue}} \\ 
    \cline{3-8}
     &   & R@2 & R@5 & R@2 & R@5 & R@2 & R@5 \\  
    \midrule
    \textbf{M\_Indexing} & $CR_{ref_{3}}$ & \underline{61.5} & 73.8 
    & 54.6 & 59.7 & 39.3 & 52.8 \\ 
    \midrule
    \multirow{3}{*}{\textbf{K\_Alignment}}
    & $CR_{ref_{3}}$ & 56.3 & 83.0 & \textbf{66.3} & 88.1 & \underline{40.0} & \underline{62.3} \\ 
    & $LFS_{ref_{1}}$ & / & / & / & / & / & / \\
    & $LFS_{ref_{3}}$ & / & / & / & / & / & / \\
    & $LFSH_{ref_{1}}$ & 55.1 & \underline{85.0} & \underline{65.9} & \textbf{92.4} & 36.1 & 58.4 \\ 
    & $LFSH_{ref_{3}}$ & \textbf{72.7} & \textbf{88.8} & 65.4 & \underline{91.9} & \textbf{48.4} & \textbf{65.6} \\ 
    \bottomrule
\end{tabular}
\caption{The recall performance of different methods across three datasets is presented. The answers to some sub-questions in the $LFS_{ref_{n}}$ method use KG reasoning without recalling supporting chunks, which is not comparable to other methods in terms of recall rate. BackBone model is DeepSeek-V2 API.}
\label{tab:AblationPerformanceOfQAReca}
\end{table*}

\begin{figure*}[htbp]
    \centering
    \includegraphics[width=0.6\linewidth]{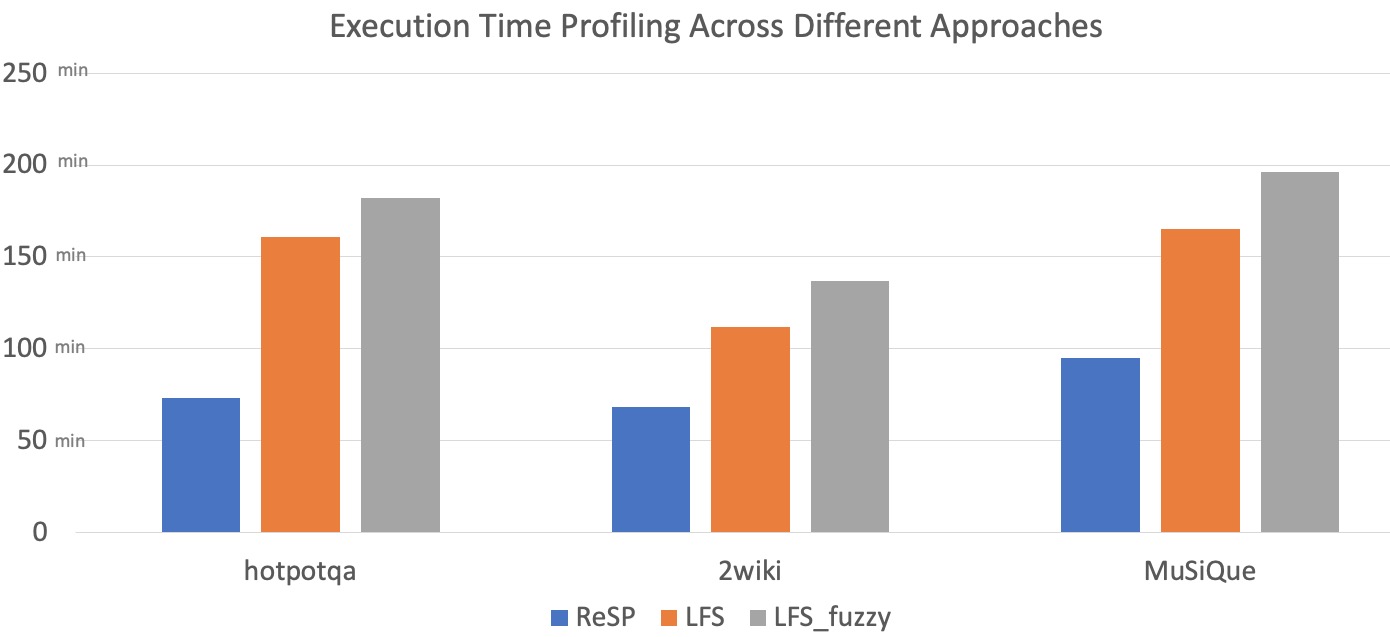}
    \caption{Each of the three test datasets comprises 1000 test problems, with 20 tasks processed concurrently and maximum number of iterations $n$ is \textbf{3}. $CR_{ref_{3}}$ method exhibits the fastest execution, whereas $LFSH_{ref_{3}}$ method is the slowest. Specifically, $CR_{ref_{3}}$ method outperforms $LFSH_{ref_{3}}$ method by 149\%, 101\%, and 134\% across the three datasets. In comparison, on the same dataset, the $LFS_{ref_{3}}$ method outperforms the $LFSH_{ref_{3}}$ method by 13\%, 22\%, and 18\%, respectively, with F1 relative losses of 2.6\%, 0.1\%, and 1.0\%, respectively. }
    \label{fig:expr_execution}
\end{figure*}

\begin{figure*}[htbp]
    \centering
    \includegraphics[width=0.8\linewidth]{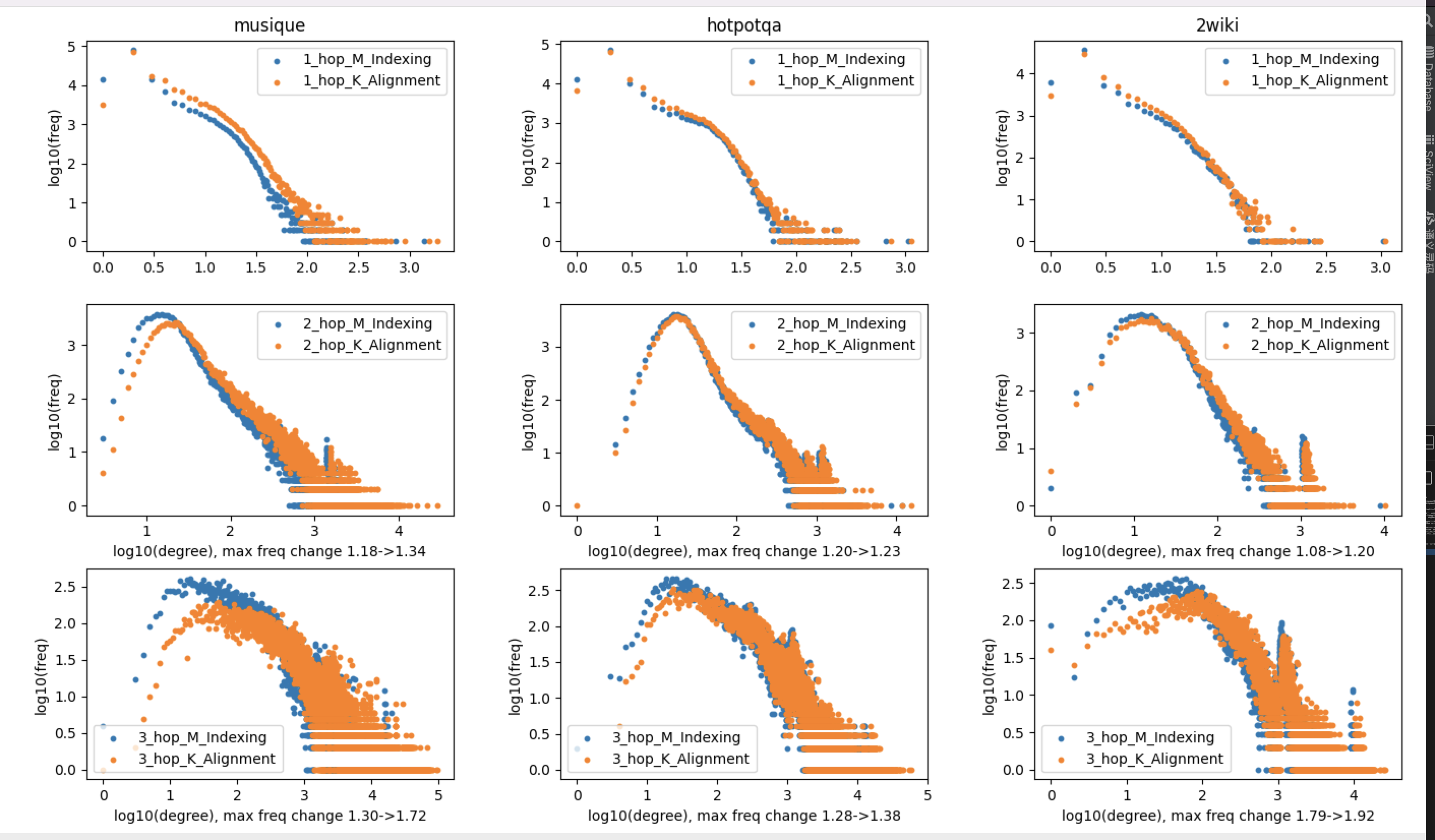}
    \caption{The connectivity of the graph exhibits a notable rightward shift after applying \textbf{K\_Alignment},the distribution changes of 1-hop, 2-hop, and 3-hop neighbors are shown.}
    \label{fig:k_alignment_vs_m_indexing}
\end{figure*}

The analysis of the experimental outcomes can be approached from the following two perspectives:

\textbf{1) Knowledge Graph Indexing.} As is shown in Table \ref{tab:AblationPerformanceOfQAReca}, after incorporation Knowledge Alignment into the KG mutual-indexing, the top-5 recall rates of $CR_{ref_3}$ improved by 9.2\%, 28.4\%, and 9.5\% respectively, with an average improvement of 15.7\%. As shown in Figure \ref{fig:k_alignment_vs_m_indexing}, after enhancing knowledge alignment, the relation density is significantly increased, and the frequency-outdegree graph is shifted to the right as a whole 
\begin{itemize}
    \item The 1-hop graph exhibits a notable rightward shift, indicating that the addition of semantic structuring has increased the number of neighbors for each node, thereby enhancing the graph's density.
    \item  The 2-hop and 3-hop graphs display an uneven distribution, with sparse regions on the left and denser regions on the right. When comparing before and after \textbf{K\_Alignment}, it is evident that the vertices in each dataset have shifted rightward, with the left side becoming more sparse. This suggests that nodes with fewer multi-hop neighbors have gained  new neighbors, leading to this observed pattern.
\end{itemize}
    
This signifies that the newly added semantic relations effectively enhance graph connectivity, thereby improving document recall rates.

\textbf{2) Graph Inference Analysis.} In terms of recall, $LFSH_{ref_{3}}$ achieves improvements over $CR_{ref_{3}}$ under the same graph index, with increases in top-5 recall rates by 15\%, 32.2\%, and 12.7\%, averaging an improvement of 19.9\%. This enhancement can be attributed to two main factors:
\begin{itemize}
    \item $LFSH_{ref_{3}}$ decomposes queries into multiple executable steps, with each sub-query retrieving chunks individually. As shown in the time analysis in Figure 8, both $LFSH_{ref_{3}}$ and $LFS_{ref_{3}}$ consume more than twice the time of $LFSH_{ref_{3}}$, indicating that increased computational time is a trade-off for improved recall rates.
    \item $LFSH_{ref_{3}}$ not only retrieves chunks but also integrates SPO triples from execution into chunk computation. Compared to $LFSH_{ref_{3}}$, it retrieves additional query-related relationships.
\end{itemize}

Due to the subgraph-based query answering in $LFS_{ref_{3}}$, it cannot be compared directly in recall rate analysis but can be examined using the F1 metric. In comparison to $LFSH_{ref_{3}}$, $LFS_{ref_{3}}$ answered questions based on the retrieved subgraphs with proportions of 33\%, 34\%, and 18\%,respectively. $LFS_{ref_{3}}$ shows a decrease in the F1 metric by 2.2\%, 0.1\%, and 0.5\%, while the computation time reduces by 12\%, 22\%, and 18\%.

The analysis of the cases with decreased performance reveals that errors or incomplete SPOs during the construction phase lead to incorrect sub-query answers, resulting in wrong final answers. This will be detailed in the case study. The reduction in computation time is primarily due to the more efficient retrieval of SPOs compared to document chunks.

In industrial applications, computation time is a crucial metric. Although $LFS_{ref_{n}}$ may introduce some errors, these can be improved through graph correction and completion. It is noteworthy that in the current experiments, the slight decrease in metrics has been traded off for reduced computation time, which we consider a feasible direction for industrial implementation.

For analyze the impact of the maximum number of iterations parameter $n$ on the results, $LFS_{ref_{1}}$ compared to $LFS_{ref_{3}}$ , the F1 scores decreased by 0.6\%, 1.6\%, and 4.8\%, respectively. Based on the experiments of $LFS_{ref_{3}}$, the proportions for an iteration count of 1 were analyzed to be 97.2\%, 94.8\%, and 87.9\%; $LFSH_{ref_{1}}$ compared to $LFSH_{ref_{3}}$ , the F1 scores decreased by 0.2\%, 1.2\%, and 4.4\%, respectively. Based on the experiments of $LFSH_{ref_{3}}$, the proportions for an iteration count of 1 were analyzed to be 98.3\%, 95.2\%, and 84.1\%; showing a positive correlation with the F1 score reduction. Table \ref{tab:case-for-lf-reasoning} provides a detailed analysis of the effect of iteration rounds on the solution of the final answer. Increasing the maximum number of iterations parameter facilitates the re-planning of existing information when $LFS_{ref_{n}}$ is unable to complete the solution, thereby addressing some unsolvable case.

\section{Applications}
\subsection{KAG for E-Goverment}

We used the KAG framework and combined it with the Alipay E-government service scenario to build a Q\&A application that supports answering users' questions about service methods, required materials, service conditions, and service locations. To build the e-government Q\&A application, we first collected 11,000 documents about government services, and based on the methods described in section 2, implemented functional modules such as index building, logical-form-guided reasoning and solving, semantic enhancement, and conditional summary generation.

During the offline index construction phase, the semantic chunking strategy is used to segment government service documents to obtain specific matters and their properties such as the administrative region, service process, required materials, service location, target audience, and the corresponding chunks.

In the reasoning and solving phase, a logical function is generated based on the given user question and graph index structure, and the logical form is executed according to the steps of the logical function. First, the index item of the administrative area where the user is located is accurately located. Then, the item name, group of people, etc. are used for search. Finally, the corresponding chunk is found through the \textit{required materials} or \textit{service process}. specifically inquired by the user.

In the semantic enhancement phase, we added two semantic relations, \textit{synonymy and hypernymy}, between items. A synonymous relation refers to items in two different regions with different names but the same meaning, such as \textit{renewal of social security card} and \textit{application for lost social security card}; a co-hypernymy relation refers to two items belonging to different subcategories under the same major category of items, such as \textit{applying for housing provident fund loan for construction of new housing} and \textit{applying for housing provident fund loan for construction and renovation of new housing}, the two items have a common hypernymy \textit{applying for housing provident fund loan}.

We compared the effects of the two technical solutions, NaiveRAG and KAG, as shown in the table below. It is evident that KAG shows significant improvements in both completeness and accuracy compared to NaiveRAG.

\renewcommand\arraystretch{1.1}
\begin{table}[htbp]
\centering
\small
\setlength\aboverulesep{0pt}\setlength\belowrulesep{0pt}
\begin{tabular*}{\textwidth}{@{\extracolsep{\fill}} l|ccc}
\toprule
     Methods & SampleNum  & Precision & Recall   \\ 
     \midrule
    NaiveRAG       & 492          & 66.5          & 52.6  \\
    KAG       & 492          & 91.6          & 71.8  \\
 \bottomrule
 \end{tabular*}
\caption{Ablation Experiments of KAG in E-Goverment Q\&A.} 
\label{Tab.e_goverment} 
\end{table}

\subsection{KAG for E-Health}
We have developed a medical Q\&A application based on the Alipay Health Manager scenario, which supports answering user's questions regarding popular science about disease, symptom, vaccine, operation, examination and laboratory test, also interpretation of medical indicators, medical recommendation, medical insurance policy inquires, hospital inquires, and doctor information inquires. We have sorted out authoritative medical document materials through a team of medical experts, and produced more than 1.8 million entities and more than 400,000 term sets, with a total of more than 5 million relations. Based on this high-quality KG, we have also produced more than 700 DSL\footnote{DSL: https://openspg.yuque.com/ndx6g9/ooil9x/sdtg4q3bw4ka5wmz} rules for indicator calculations to answer the questions of indicator interpretation.\newline \newline
During the knowledge construction phase, a strongly constrained schema is used to achieve precise structural definition of entities such as diseases, symptoms, medications, and medical examinations. This approach facilitates accurate answers to questions and generates accurate knowledge, while also ensuring the rigor of relations between entities. 
In the reasoning phase, the logical form is generated based on the user's query, and then translated to DSL form for the query on KG. The query result is returned in the form of triples as the answer.
The logical form not only indicates how to query the KG, but also contains the key structural information in the user’s query (such as city, gender, age, indicator value, etc.). When parsing the logical form for query in graph, the DSL rules which produced by medical expert will also be triggered, and the conclusion will be returned in the form of triples. For example, if a user asks about \textit{"blood pressure 160"}, it will trigger the rules as:\begin{figure}[H]
    \raggedright
    \includegraphics[width=0.7\linewidth]{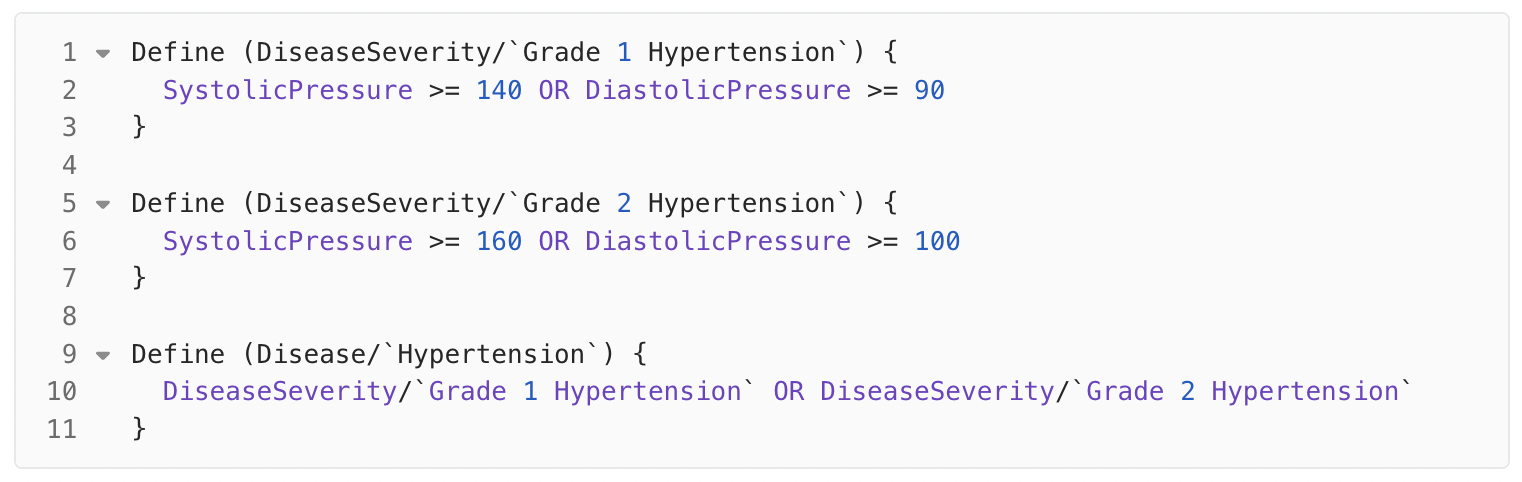}
\end{figure}    
, which strictly follows the defination of $\mathcal{L}$ in LLMFriSPG, and the conclusion that the person may have hypertension will be obtained.

In the semantic enhancement phase, we utilize the term set to express the two semantic relations of synonymy and hypernym of concepts. The hypernym supports the expression of multiple hypernyms. During knowledge construction and user Q\&A phase, entities are aligned with medical terms. For example, in the concept of surgery type, the hypernym of deciduous tooth extraction and anterior tooth extraction is tooth extraction. When the user only asks questions about tooth extraction, all its hyponyms can be retrieved based on the term, and then the related entity information can be retrieved for answering.
With the support of KAG, we achieved a recall rate of 60.67\% and a precision rate of 81.32\% on the evaluation set which sampling online Q\&A queries. In the end-to-end scenario, the accuracy of medical insurance policy inquires (Beijing, Shanghai, Hangzhou) reached 77.2\%, and the accuracy rate of popular science intentions has exceeded 94\%, and the accuracy rate of interpreting indicator intentions has exceeded 93\%.
\section{Related Works}

\subsection{DIKW Pyramid}
Following the DIKW pyramid theories\cite{ackoff1989data, baskarada2013data, terra2003understanding, hey2004data}, after data is processed and contextualised, it becomes information, and by integrating information with experience, understanding, and expertise, we gain knowledge. We usually use information extraction technology to obtain information from the original text\cite{sarawagi2008information,weikum2010information,piskorski2013information}, and obtain knowledge from the information through linking, fusion, analysis, and learning technology\cite{terra2003understanding, sajja2010knowledge, weikum2010information}. Information and knowledge are a single entity having different forms. There are no unified language to represent data, information and knowledge, RDF/OWL\cite{allemang2011semantic} only provides binary representation in the form of triples, and LPG\cite{sharma2019schema} lacks support for knowledge semantics and classification. SPG\footnote{Official site of SPG: https://spg.openkg.cn/en-US}\cite{yikgfabric} supports knowledge hierarchy and classification representation, but lacks text context support that is friendly to large language models. Our proposed LLMFriSPG supports hierarchical representation from data to information to knowledge, and also provides reverse context-enhanced mutual-indexing. 
\subsection{Vector Similarity-based RAG}
The external knowledge base use the traditional search engine provides an effective method for updating the knowledge of LLMs, it retrievals supporting documents by calculating the text or vector similarity\cite{gao2023retrieval, fan2024survey} between the query and document, and then answers questions using the in-context learning method of LLMs. In addition, this method faces great challenges in understanding long-distance knowledge associations between documents. Simple vector-based retrieval is not suitable for multi-step reasoning or tracking logical links between different information fragments. To address these challenges, researchers have explored methods such as fine-grained document segmentation, CoT\cite{trivedi2022interleaving}, and interactive retrieval\cite{jiang2024retrieve,shao2023enhancing}. Despite these optimizations, traditional query-chunks similarity methods still has difficulty in accurately focusing on the relations between key knowledge in complex questions, resulting in low information density and ineffective association of remote knowledge. We will illustrate the logical-form-guided solving method.

\subsection{Information Retrieval-based GraphRAG}
This type of methods use information extraction techniques to build entity and relation associations between different documents, which can better perceive the global information of all documents. Typical tasks in the knowledge construction phase include: graph information extraction and knowledge construction\&enhancement. Methods like GraphRAG\cite{edge2024local}, ToG 2.0\cite{ma2024think}, HippoRAG\cite{gutierrez2024hipporag} use OpenIE to extract graph-structure information like entities and relations, some of them exploit multi-hop associations between entities to improve the effectiveness of cross-document retrieval\cite{ma2024think, gutierrez2024hipporag}, methods like DALK\cite{li2024dalk} use PubTator Central(PTC) annotation to reduce the noise problem of openIE, some of them utilize entity disambiguation technology to enhance the consistency of graph information\cite{gutierrez2024hipporag,sarmah2024hybridrag}. GraphRAG\cite{edge2024local} generates element-level and community-level summaries when building offline indexes, and it uses a QFS\cite{dang2006duc} method to first calculate the partial response of each summary to the query and then calculate the final response. This inherent characteristic of GraphRAG's hierarchical summarization makes it difficult to solve questions such as multi-hop Q\&A and incremental updates of documents. KGs constructed by openIE contains a lot of noise or irrelevant information\cite{zhang2020aser,bi2024codekgc,fang2021discos}. According to the DIKW pyramid hierarchy, these methods only extract the information graph structure and make limited attempts to disambiguate entities in the transformation of information into knowledge,but they do not address issues such as semantic directionality and logical sensitivity. This paper will introduce a method in KAG to enhance information-to-knowledge conversion based on domain concept semantic graph alignment.

\subsection{KG-based Question and Answering}
Reasoning based on traditional KGs has good explainability and transparency, but is limited by the scale of the domain KG, the comprehensiveness of knowledge, the detailed knowledge coverage, and the timeliness of updates\cite{tian2022knowledge}. n this paper, we introduce HybridReasoning to alleviate issues such as knowledge sparsity, inconsistent entity granularity, and high graph construction costs. The approach leverages KG retrieval and reasoning to enhance generation, rather than completely replacing RAG.

To achieve KG-enhanced generation, it is necessary to address KG-based knowledge retrieval and reasoning. One approach is knowledge edge retrieval (IR)\cite{yao2007knowledge}, which narrows down the scope by locating the most relevant entities, relations, or triples based on the question. Another approach is semantic parsing (SP)\cite{berant2013semantic,guo2018dialog}, which converts the question from unstructured natural language descriptions into executable database query languages (such as SQL, SPARQL\cite{perez2006semantics}, DSL\footnote{DSL: https://openspg.yuque.com/ndx6g9/ooil9x/sdtg4q3bw4ka5wmz}, etc.), or first generates structured logical forms (such as S-expressions\cite{gu2021beyond,luo2024chatkbqa}) and then converts them into query languages.

Although conversational QA over large-scale knowledge bases can be achieved without explicit semantic parsing (e.g., HRED-KVM\cite{kacupaj2021conversational}), most work focuses on exploring context-aware semantic parsers\cite{guo2018dialog,lan2021modeling,luo2024chatkbqa}.

Some papers use sequence-to-sequence models to directly generate query languages\cite{kapanipathi2021leveraging,omar2023universal}. These methods are developed for a specific query language, and sometimes even for a specific dataset, lacking generality for supporting different types of structured data. Others use step-by-step query graph generation and search strategies for semantic parsing\cite{atif2023beamqa,jiang2023structgpt,gu2023dont}. This method is prone to uncontrollable issues generated by LLM, making queries difficult and having poor interpretability. Methods like ChatKBQA\cite{luo2024chatkbqa}, CBR-KBQA\cite{das2021case} completely generate S-expressions and provide various enhancements for the semantic parsing process. However, the structure of S-expressions is relatively complex, and integrating multi-hop questions makes it difficult for LLMs to understand and inconvenient for integrating KBQA and RAG for comprehensive retrieval. To address these issues, we propose a multi-step decomposed logical form to express the multi-hop retrieval and reasoning process, breaking down complex queries into multiple sub-queries and providing corresponding logical expressions, thereby achieving integrated retrieval of SPO and chunks.

\subsection{Bidirectional-enhancement of LLMs and KGs}
LLM and KG are two typical neural and symbolic knowledge utilization methods. Since the pre-trained language model such as BERT \cite{DBLP:conf/naacl/DevlinCLT19}, well-performed language models are used to help improve the tasks of KGs. The LLMs with strong generalization capability are especially believed to be helpful in the life-cycle of KGs. There are a lot of works conducted to explore the potential of LLMs for in-KG and out-of-KG tasks. For example, using LLMs to generate triples to complete triples is proved to be much cheaper than the traditional human-centric KG construction process, with acceptable accuracy for the popular entities \cite{DBLP:conf/emnlp/VeseliRKW23}.
In the past decade, methods for in-KG tasks are designed by learning from KG structures, such as structure embedding-based methods. The text information such as names and descriptions of entities is not fully utilized due to the limited text understanding capability of natural language processing methods until LLMs provide a way. Some works using LLMs for text semantic understanding and reasoning of entities and relations in KG completion 
 \cite{KoPA}, rule learning \cite{chatrule}, complex logic querying \cite{DBLP:journals/corr/abs-2305-01157}, etc. On the other way, KGs are also widely used to improve the performance of LLMs. For example, using KGs as external resources to provide accurate factual information, mitigating hallucination of LLMs during answer generation \cite{ma2024think}, generating complex logical questions answering planning data to fine-tune the LLMs, improving LLMs planning capability and finally improving its logical reasoning capability \cite{DBLP:journals/corr/abs-2406-14282}, using KGs to uncover associated knowledge that has changed due to editing for better knowledge editing of LLMs \cite{DBLP:journals/corr/abs-2402-13593}, etc. The bidirectional-enhancement of LLMs and KGs is widely explored and partially achieved. 

\section{Limitations}
In this article, we have proven the adaptability of the KAG framework in Q\&A scenarios in vertical and open domains. However, the currently developed version of OpenSPG-KAG 0.5 still has major limitations that need to be continuously overcome, such as:

\textbf{Implementing our framework requires multiple LLM calls during the construction and solving phases.} A substantial number of intermediate tokens required to be generated during the planning stage to facilitate the breakdown of sub-problems and symbolic representation, this leads to computational and economic overhead, as illustrated in Table \ref{tab:case-for-reasoning}, where the problem decomposition not only outputs sub-problems but also logical functions, resulting in approximately twice as many generated tokens compared to merely decomposing the sub-problems. Meanwhile, currently, all model invocations within the KAG framework, including entity recognition, relation extraction, relation recall, and standardization, rely on large models. This multitude of models significantly increases the overall runtime. In future domain-specific implementations, tasks like relation recall, entity recognition, and standardization could be substituted with smaller, domain-specific models to enhance operational efficiency.

\textbf{The ability to decompose and plan for complex problems requires a high level of capability}. Currently, this is implemented using LLMs, but planning for complex issues remains a significant challenge. For instance, when the task is to compare who is older, the problem should be decomposed into comparing who was born earlier. Directly asking for age is not appropriate, as they are deceased, and \text"what is the age" refers to the age at death, which doesn't indicate who is older. 
Decomposing and planning complex problems necessitates ensuring the model's accuracy, stability, and solvability in problem decomposition and planning. The current version of the KAG framework does not yet address optimizations in these areas.
We will further explore how pre-training, SFT, and COT strategies can improve the model's adaptability to logical forms and its planning and reasoning capabilities.

\begin{quote}
\small
    Question: \textit{Which film has the director who is older, God'S Gift To Women or Aldri Annet Enn Bråk?} \\
    Q1: \textit{Which director directed the film God'S Gift To Women?}  A1: \textbf{Michael Curtiz} \\
    Q2: \textit{Which director directed the film Aldri Annet Enn Bråk?}  A2: \textbf{Edith Carlmar} \\
    Q3: \textit{What is the age of the director of God'S Gift To Women? } A3: \textbf{74 years old.} \textit{Michael Curtiz (December 24, 1886 to April 11, 1962)...}\\
    Q4: \textit{What is the age of the director of Aldri Annet Enn Bråk?}  A4: \textbf{91 years old.} \textit{Edith Carlmar (Edith Mary Johanne Mathiesen)  (15 November 1911\ to 17 May 2003) ...}\\
    Q5: \textit{Compare the ages of the two directors to determine which one is older.}  A5: \textcolor{red}{\textbf{Edith Carlmar is older}.} \textit{Actually, Michael Curtiz was born earlier}.
\end{quote}

\textbf{OpenIE significantly lowers the threshold for building KGs, but it also obviously increases the technical challenges of knowledge alignment.} Although the experiments in this article have shown that the accuracy and connectivity of extracted knowledge can be improved through knowledge alignment. However, there are still more technical challenges waiting to be overcome, such as optimizing the accuracy of multiple-knowledge(such as events, rules, pipeline, etc.) extraction and the consistency of multiple rounds of extraction. In addition, schema-constraint knowledge extraction based on the experience of domain experts is also a key way to obtain rigorous domain knowledge, although the labor cost is high. These two methods should be applied collaboratively to better balance the requirements of vertical scenarios for the rigor of complex decision-making and the convenience of information retrieval. For instance, when extracting team members from multiple texts and asked about the total number of team members, a comprehensive extraction is crucial for providing an accurate answer based on the structured search results. Incorrect extractions also impair response accuracy.

\section{Conclusion and Future Work}
In order to build professional knowledge services in vertical domains, fully activate the capabilities and advantages of symbolic KGs and parameterized LLMs, and at the same time significantly reduce the construction cost of domain KGs, we proposed the KAG framework and try to accelerated its application in professional domains.
In this article, we introduce in detail the knowledge accuracy, information completeness and logical rigorous are the key characteristics that professional knowledge services must have. At the same time, we also introduce innovations such as LLMs friendly knowledge representation, mutual-indexing of knowledge structure and text chunks, knowledge alignment by semantic reasoning, logic-form-guided hybrid reasoning\&solving and KAG model. Compared with the current most competitive SOTA method, KAG has achieved significant improvements on public data sets such as HotpotQA, 2wiki, musique. We have also conducted case verifications in E-goverment Q\&A and E-Health Q\&A scenarios of Alipay, further proving the adaptability of the KAG framework in professional domains.\newline 

In the future, there is still more work to be explored to continuously reduce the cost of KG construction and improve the interpretability and transparency of reasoning, such as multiple knowledge extraction, knowledge alignment based on \textbf{OneGraph}, domain knowledge injection, large-scale instruction synthesis, illusion suppression of knowledge logic constraints, etc. \newline

This study does not encompass the enhancement of models for decomposing and planning complex problems, which remains a significant area for future research. In future work, KAG can be employed as a reward model to provide feedback and assess the model's accuracy, stability, and solvability through the execution of planning results, thereby enhancing the capabilities of planning models. \newline

We will also work in depth with the community organization \textbf{OpenKG} to continue to tackle key technical issues in the collaboration between LLMs and KGs.

\section{Acknowledgements}
This work was completed by the AntGroup Knowledge Graph Team, in addition to the authors in the list, other contributors include Yuxiao He, Deng Zhao, Xiaodong Yan, Dong Han, Fanzhuang Meng, Yang Lv, Zhiying Yin, etc, thank you all for your continuous innovation attempts and hard work. This work also received strong support from Professor Huajun Chen, Researcher Wen Zhang of Zhejiang University, and Professor Wenguang Chen of AntGroup Technology Research Institute, thank you all.

\bibliographystyle{unsrt}
\bibliography{bibfile} 

\clearpage
\appendix
\section{Example of KAG Solver}

\begin{table*}[htbp]
    \centering
    \scriptsize
\begin{tabular}{l}
	\toprule
	\begin{tabular}{p{14cm}} 
		\textbf{Round One}
	\end{tabular} \\ \midrule
        \begin{tabular}{p{14cm}} 
            \textbf{Initial Question: How many times did the plague occur in the birth place of Concerto in C Major Op 3 6's composer?} \\ \\
            \textbf{Step1}: \textcolor{gray}{Who is the composer of Concerto in C Major Op 3 6?.} \\
            \textbf{Logical Form}: \textcolor{blue}{Retrieval}(s=s1:Work[C Major Op 3 6],p=p1:\textcolor{brown}{composer},o=o1:\textcolor{brown}{Person})\\
            \textbf{Graph Retrieval}: (antonio vivaldi, composed concerto in, c major op 3 6)\\
            \textbf{Hybird Retrieval}: not executed \\
            \textbf{A1}: Antonio Vivaldi \\ \\
            
            \textbf{Step2}: \textcolor{gray}{What is the birthplace of this composer?}\\
            \textbf{Logical Form}: \textcolor{blue}{Retrieval}(s=o1,p=p2:\textcolor{brown}{brithPlace},o=o2:\textcolor{brown}{Place})\\
            \textbf{Graph Retrieval}: not found\\
            \textbf{Hybird Retrieval}: \\
            doc1:Antonio Vivaldi composed a mandolin concerto (Concerto in C major Op.3 6) ... \\
            doc2:Antonio Lucio Vivaldi (Italian: ... Born in \textbf{Venice}... \\
            ...\\
            \textbf{A2}: Venice \\ \\

            \textbf{Step3}: \textcolor{gray}{How many times did the plague occur in this birthplace?}\\
            \textbf{Logical Form}: \textcolor{blue}{Retrieval}(s=o2,p=p3:\textcolor{brown}{plagueOccurrence},o=o3:\textcolor{brown}{Plague})\\
            \textbf{Graph Retrieval}: not found\\
            \textbf{Hybird Retrieval}: \\
            doc1:Antonio Vivaldi composed a mandolin concerto (Concerto in C major Op.3 6) ... \\
            doc2:The Black Death, also known as the Great Plague.. in Europe from 1347 to 1351.. \\
            ...\\
            \textbf{A3}: I don't know \\ \\

            \textbf{Step4}: \textcolor{gray}{Count the number of plague occurrences.}\\
            \textbf{Logical Form}: \textcolor{blue}{Count}(o3)->nums\\
            \textbf{Graph Retrieval}: not found\\
            \textbf{Hybird Retrieval}: not found\\
            \textbf{A4}: I don't know\\
            
	\end{tabular} \\ \midrule

        \begin{tabular}{p{14cm}} 
		\textbf{Round Two}
	\end{tabular} \\ \midrule
        \begin{tabular}{p{14cm}} 
            \textbf{Rewrite Question: What specific records or historical accounts detail the number of plague occurrences in Venice during the mid-14th century?} \\ \\
            \textbf{Step1}: \textcolor{gray}{What specific records or historical accounts detail the number of plague occurrences in Venice during the mid-14th century?} \\
            \textbf{Logical Form}: \textcolor{blue}{Retrieval}(s=s1:\textcolor{brown}{City}[Venice],p=p3:\textcolor{brown}{plagueOccurrencesInMid14thCentury},o=o1:\textcolor{brown}{Times})\\
            \textbf{Graph Retrieval}: not found\\
            \textbf{Hybird Retrieval}:\\
            doc1:In 1466, perhaps 40,000 people died of the plague...\textbf{Plague occurred in Venice 22 times between 1361 and 1528}...\\
            doc2:The Black Death, also known as the Great Plague...\\
            ...\\
            \textbf{A1}: The plague occurred in Venice 22 times between 1361 and 1528. The 1576–77 plague killed 50,000, almost a third of the population. \\ 
	\end{tabular} \\ \midrule

        \begin{tabular}{p{14cm}} 
		\textbf{Final Answer}
	\end{tabular} \\ \midrule
        \begin{tabular}{p{14cm}} 
            \textbf{Question: How many times did the plague occur in the birth place of Concerto in C Major Op 3 6's composer?} \\ \\
            \textbf{Step1}: What specific records or historical accounts detail the number of plague occurrences in Venice during the mid-14th century? \\
            \textbf{A}: 22 times \\   
	\end{tabular} \\ \midrule
        
\end{tabular}
\caption{An example of using logical-from to guide question planning, reasoning, retrieval, and answer generation, and using multiple rounds of reflection to rephrase questions.}
\label{tab:case-for-lf-reasoning}
\end{table*}

\section{Example of Logical form Reasoner}

\renewcommand\arraystretch{1.1}
\begin{table*}[htbp]
    \centering
    \scriptsize
\begin{tabular}{l}
	\toprule
	\begin{tabular}{p{14cm}} 
		\textbf{Numerical Reasoning}
	\end{tabular} \\ \midrule
        \begin{tabular}{p{14cm}} 
            \textbf{question: Which sports team for which Cristiano Ronaldo played in 2011 was founded last ?} \\
            \textbf{Step1}: \textcolor{gray}{Identify the Sports Teams Cristiano Ronaldo Played for in 2011 .} \\
            Logical Form: \textcolor{blue}{Retrieval(}s=s1:\textcolor{brown}{Player}[Cristiano Ronaldo], p=p1:\textcolor{brown}{playedFor}, o=o1:\textcolor{brown}{SportsTeam}, p.PlayedForInYear=2011) \\
            \textbf{Step2}: \textcolor{gray}{Determine the Foundation Years of Each Identified Team.} \\
            \textbf{Logical Form}: \textcolor{blue}{Retrieval}(s=o1, p=p2:\textcolor{brown}{foundationYear}, o=o2:\textcolor{brown}{Year}) \\
            \textbf{Step3}: \textcolor{gray}{Which team was founded last?} \\
            \textbf{Logical Form}: \textcolor{blue}{Sort}(set=o1, orderby=o2, direction=max, limit=1) 
	\end{tabular} \\  
        \begin{tabular}{p{14cm}} 
            \\
            \textbf{question: What is the sum of 30 + 6 and the age of the founder of Tesla in 2027 ?} \\
            \textbf{Step1}: \textcolor{gray}{What is the sum of 30 + 6 ?} \\
            \textbf{Logical Form}:  math1 = \textcolor{blue}{Math}(30+6) \\
            \textbf{Step2}:  \textcolor{gray}{Who is the founder of Tesla?} \\
            \textbf{Logical Form}:  \textcolor{blue}{Retrieval}(s=s2:\textcolor{brown}{Company}[Tesla], p=p2:\textcolor{brown}{founder}, o=o2) \\
            \textbf{Step3}: \textcolor{gray}{In which year was the founder of Tesla born?} \\
            \textbf{Logical Form}:  \textcolor{blue}{Retrieval}(s=o2, p=p3:\textcolor{brown}{yearOfBirth}, o=o3) \\
            \textbf{Step4}: \textcolor{gray}{How old will the founder of Tesla be in the year 2027?} \\
            \textbf{Logical Form}:  math4 = \textcolor{blue}{Math}(2027-o3) \\
            \textbf{Step5}: \textcolor{gray}{What is the sum of math1 and math4?} \\
            \textbf{Logical Form}:  math5 = \textcolor{blue}{Math}(math1+math4) 
	\end{tabular} \\  \midrule
        \begin{tabular}{p{14cm}} 
		\textbf{Logical Reasoning}
	\end{tabular} \\ \midrule
        \begin{tabular}{p{14cm}} 
            \textbf{question: Find a picture containing vegetables or fruits.} \\
            \textbf{Step1}: \textcolor{gray}{Find pictures containing vegetables.} \\
            \textbf{Logical Form}: \textcolor{blue}{Retrieval}(s=s1:\textcolor{brown}{Image}, p=p2:\textcolor{brown}{contains}, o=o1:\textcolor{brown}{Vegetables}) \\
            \textbf{Step2}: \textcolor{gray}{Find pictures containing fruits.} \\
            \textbf{Action2}: \textcolor{blue}{Retrieval}(s=s2:\textcolor{brown}{Image}, p=p2:\textcolor{brown}{contains}, o=o2:\textcolor{brown}{Fruits}) \\
            \textbf{Step3}: Output s1, s2. \\
            \textbf{Logical Form}:  \textcolor{blue}{Output}(s1, s2)
	\end{tabular} \\  
        \begin{tabular}{p{14cm}} 
            \\
            \textbf{question: Find a picture containing vegetables and fruits.} \\
            \textbf{Step1}: \textcolor{gray}{Find pictures containing vegetables.} \\
            \textbf{Logical Form}: \textcolor{blue}{Retrieval}(s=s1:\textcolor{brown}{Image}, p=p2:\textcolor{brown}{contains}, o=o1:\textcolor{brown}{Vegetables}) \\
            \textbf{Step2}: \textcolor{gray}{Find pictures containing fruits.} \\
            \textbf{Logical Form}: \textcolor{blue}{Retrieval}(s=s1, p=p2:\textcolor{brown}{contains}, o=o2:\textcolor{brown}{Fruits}) \\
            Step3: Output s1. \\
            \textbf{Logical Form}:  \textcolor{blue}{Output}(s1)
	\end{tabular} \\  \midrule
        \begin{tabular}{p{14cm}} 
		\textbf{Semantic Deduce}
	\end{tabular} \\ \midrule
        \begin{tabular}{p{14cm}} 
            \textbf{question: Do I need to present the original ID card when applying for a passport?} \\
            \textbf{Step1}: \textcolor{gray}{What documents are required to apply for a passport?} \\
            \textbf{Logical Form}: \textcolor{blue}{Retrieval}(s=s1:\textcolor{brown}{Event}[apply for a passport], p=p1:\textcolor{brown}{support\_chunks}, o=o1:\textcolor{brown}{Chunk}) \\
            \textbf{Step2}: \textcolor{gray}{Does this set of documents include the original identity card?} \\
            \textbf{Logical Form}: \textcolor{blue}{Deduce}(left=o1, right=the original identity card, op=entailment)
	\end{tabular} \\  
	\bottomrule
\end{tabular}
\caption{The cases of reasoning with logical form}
\label{tab:case-for-reasoning}
\end{table*}

\end{document}